\documentclass[runningheads]{llncs}
\usepackage{graphicx}
\usepackage{tikz}
\usepackage{comment}
\usepackage{amsmath,amssymb} % define this before the line numbering.
\usepackage{color}
\usepackage{glossaries}
\usepackage{booktabs}               
\usepackage{multirow}
\usepackage{bbm}
\usepackage{subcaption}
\usepackage{wrapfig,booktabs}
\usepackage{xcolor}
\definecolor{gain}{HTML}{008000}
\definecolor{loss}{HTML}{E32636}

\usepackage{epigraph}
\usepackage[accsupp]{axessibility}  % Improves PDF readability for those with disabilities.

\newglossaryentry{ResNet}
{
    name={ResNet},
    description={ResNet}
}
\newglossaryentry{OccamNet}
{
    name={OccamNet},
    description={OccamNet}
}
\newglossaryentry{OccamResNet}
{
    name={OccamResNet},
    description={OccamResNet}
}
\newglossaryentry{OccamResNets}
{
    name={OccamResNets},
    description={OccamResNets}
}

\newglossaryentry{BiasedMNIST}
{
    name={Biased MNISTv2},
    description={Biased MNISTv2}
}

\newglossaryentry{COCO}
{
    name={COCO-on-Places},
    description={Coco-on-Places}
}

\newglossaryentry{BAR}
{
    name=BAR,
    description={Biased Action Recognition}
}

\newglossaryentry{ERM}
{
    name=ERM,
    description={Empirical Risk Minimization}
}
\newglossaryentry{UpWt}
{
    name=Up Wt,
    description={Group Upweighting}
}
\newglossaryentry{SD}
{
    name=SD,
    description={Spectral Decoupling}
}
\newglossaryentry{gDRO}
{
    name=gDRO,
    description={Group DRO}
}
\newglossaryentry{PGI}
{
    name=PGI,
    description={Predictive Group Invariance}
}
\newglossaryentry{OccamNets}
{
    name={OccamNets},
    description={OccamNets}
}

\begin{document}
\newcommand{\ck}[1]{\textcolor{red}{CK: #1}}
\newcommand{\rs}[1]{\textcolor{blue}{RS: #1}}
\newcommand{\kk}[1]{\textcolor{magenta}{KK: #1}}
% \newcommand{\th}[1]{ \textcolor{green}{TH: #1}}

% \linenumbers
\pagestyle{headings}
\mainmatter
\def\ECCVSubNumber{2808}  % Insert your submission number here

\title{OccamNets: Mitigating Dataset Bias by Favoring Simpler Hypotheses}

\begin{comment}
\titlerunning{ECCV-22 submission ID \ECCVSubNumber} 
\authorrunning{ECCV-22 submission ID \ECCVSubNumber} 
\author{Anonymous ECCV submission}
\institute{Paper ID \ECCVSubNumber}
\end{comment}
%******************

%\begin{comment}
\titlerunning{OccamNets}
\author{Robik Shrestha\inst{1}\orcidID{0000-0002-0945-3458} \and
Kushal Kafle\inst{2}\orcidID{0000-0002-0847-7861} \and
Christopher Kanan\inst{1,3}\orcidID{0000-0002-6412-995X}}
\authorrunning{R. Shrestha et al.}
\institute{Rochester Institute of Technology, Rochester, USA\\ \email{rss9369@rit.edu} \and
Adobe Research, San Jose, USA\\ \email{kkafle@adobe.com} \and
University of Rochester, Rochester, USA\\  \email{ckanan@cs.rochester.edu}
}

%\institute{Princeton University, Princeton NJ 08544, USA \and
%Springer Heidelberg, Tiergartenstr. 17, 69121 Heidelberg, Germany
%\end{comment}
%******************
\maketitle
% \rs{TODO: Hone on these points more}
% \begin{itemize}
%     \item Do not rely on oracle bias labels
%     \item Show good results on both in and out-of-distribution (need to see if this is still true)
%     \item Same set of Occam-specific hyperparameters across datasets
%     \item Simpler hypotheses suffice for many samples
% \end{itemize}
\begin{abstract}
    Dataset bias and spurious correlations can significantly impair generalization in deep neural networks. Many prior efforts have addressed this problem using either alternative loss functions or sampling strategies that focus on rare patterns. We propose a new direction: modifying the network architecture to impose inductive biases that make the network robust to dataset bias. Specifically, we propose \gls*{OccamNets}, which are biased to favor simpler solutions by design. \gls*{OccamNets} have two inductive biases. First, they are biased to use as little network depth as needed for an individual example. Second, they are biased toward using fewer image locations for prediction. While \gls*{OccamNets} are biased toward simpler hypotheses, they can learn more complex hypotheses if necessary. In experiments, \gls*{OccamNets} outperform or rival state-of-the-art methods run on architectures that do not incorporate these inductive biases. {Furthermore, we demonstrate that when the state-of-the-art debiasing methods are combined with \gls*{OccamNets}\footnote{https://github.com/erobic/occam-nets-v1} results further improve}.
\end{abstract}

\section{Introduction}

% \begin{center}
% \footnotesize
% \textit{Frustra fit per plura quod potest fieri per pauciora.}\\
% \textbf{William of Occam}, \textit{Summa Totius Logicae (1323 CE)}
% %\\

% \end{center}

\epigraph{Frustra fit per plura quod potest fieri per pauciora}{\small \textbf{William of Occam}, \textit{Summa Totius Logicae (1323 CE)}}

% \epigraph{Frustra fit per plura quod potest fieri per pauciora\\
% \textit{It is futile to do with more things that which can be done with fewer} }{\small{} \textbf{William of Occam}, \textit{Summa Totius Logicae (1323 CE)}}

% \epigraph{Frustra fit per plura quod potest fieri per pauciora. (It is futile to do with more things that which can be done with fewer.) }{\textit{Mac Flecknoe \\ John Dryden}}

Spurious correlations and dataset bias greatly impair generalization in deep neural networks~\cite{agrawal2018don,bolukbasi2016man,he2009learning,shrestha2022investigation}. This problem has been heavily studied. The most common approaches are re-sampling strategies~\cite{chawla2002smote,cui2019class,he2008adasyn,sagawa2020investigation}, altering optimization to mitigate bias~\cite{DBLP:journals/corr/abs-1911-08731}, adversarial unlearning~\cite{adeli2019bias,grand2019adversarial,ramakrishnan2018overcoming,zhang2018mitigating}, learning invariant representations~\cite{arjovsky2019invariant,choe2020empirical,teney2020unshuffling}, and ensembling with bias-amplified models~\cite{cadene2019rubi,clark2019don,nam2020learning}. Here, we propose a new approach: incorporating architectural inductive biases that combat dataset bias.  %  (see Fig.~\ref{fig:overview})
% \begin{wrapfigure}[17]{4}{0.45\textwidth}
%  \centering
%     \includegraphics[width=\linewidth]{figures/overview.png}
%     \caption{With \gls*{OccamNets}, we propose exploring architectural inductive biases which is an orthogonal direction of tackling biases compared to the existing works.}
%      \label{fig:overview}
% \end{wrapfigure}
\begin{figure}[ht!]
\footnotesize
 \centering
    \includegraphics[width=0.9\linewidth]{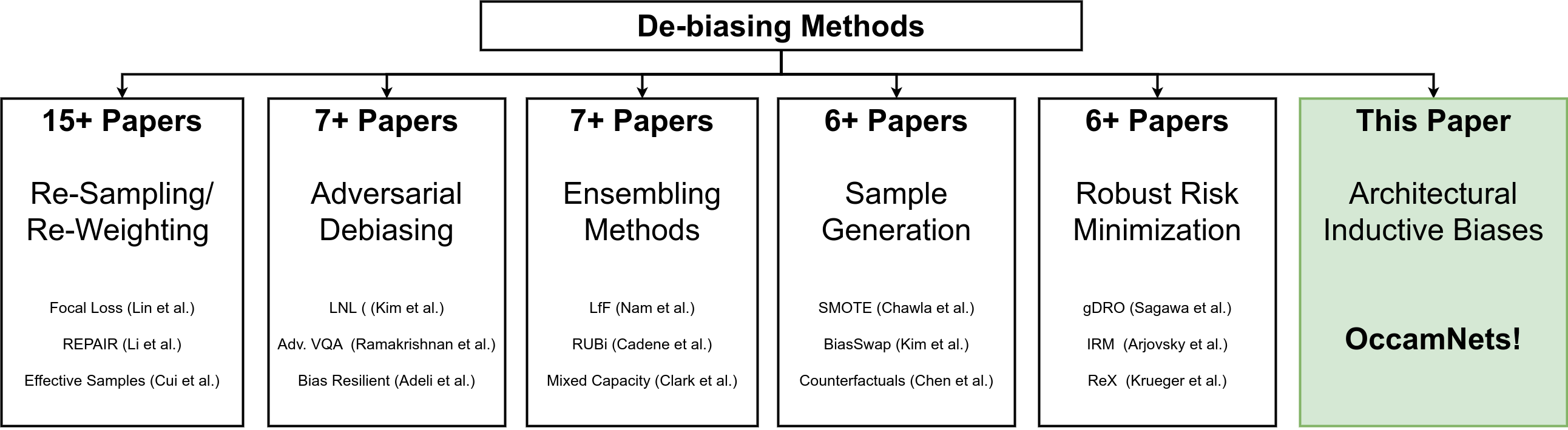}
      \caption{\gls*{OccamNets} focus on architectural inductive biases, which is an orthogonal direction to tackling dataset biases compared to the existing works.}
      \label{fig:overview}
\end{figure}

In a typical feedforward network, each layer can be considered as computing a function of the previous layer, with each additional layer making the hypothesis more complex. Given a system trained to predict multiple categories, with some being highly biased, this means the network uses the same level of complexity across all of the examples, even when some examples should be classified with simpler hypotheses (e.g., less depth). Likewise, pooling in networks is typically uniform in nature, so every location is used for prediction, rather than only the minimum amount of information. In other words, typical networks violate Occam's razor. Consider the \gls*{BiasedMNIST} dataset~\cite{shrestha2022investigation}, where the task is to recognize a digit while remaining invariant to multiple spuriously correlated factors, which include colors, textures, and contextual biases. The most complex hypothesis would exploit every factor during classification, including the digit's color, texture, or background context. A simple hypothesis would instead be to focus on the digit's shape and to ignore these spuriously correlated factors that work very well during training but do not generalize. We argue that a network should be capable of adapting its hypothesis space for each example, rather than always resorting to the most complex hypothesis, which would help it to ignore extraneous variables that hinder generalization.

Here, we propose convolutional \gls*{OccamNets} which have architectural inductive biases that favor using the minimal amount of network depth and the minimal number of image locations during inference for a given example. The first inductive bias is implemented using early exiting, which has been previously studied for speeding up inference. The network is trained such that later layers focus on examples earlier layers find hard, with a bias toward exiting early. The second inductive bias replaces global average pooling before a classification layer with a function that is regularized to favor pooling with fewer image locations from class activation maps (CAMs). We hypothesize this would be especially useful for combating background and contextual biases~\cite{ahmed2020systematic,singh2020don}. OccamNets are complementary to existing approaches and can be combined with them.

\paragraph{In this paper, we demonstrate that architectural inductive biases are effective at mitigating dataset bias. Our specific contributions are:}
\begin{itemize}
    \item We introduce the \gls*{OccamNet} architecture, which has architectural inductive biases for favoring simpler solutions to help overcome dataset biases. \gls*{OccamNets} do not require the biases to be explicitly specified during training, unlike many state-of-the-art debiasing algorithms.
    \item In experiments using biased vision datasets, we demonstrate that \gls*{OccamNets} greatly outperform architectures that do not use the proposed inductive biases. Moreover, we show that \gls*{OccamNets} outperform or rival existing debiasing methods that use conventional network architectures. 
    \item We combine \gls*{OccamNets} with four recent debiasing methods, which all show improved results compared to using them with conventional architectures. 
    %\item We show that OccamNets can be combined with existing methods for mitigating bias. \kk{We demonstrate that OccamNet can be combined with other bias mitigation algorithms by swapping their image encoder with OccamNet-enchanced variants}
    %\item  \gls*{OccamNets} outperform or rival state-of-the-art bias mitigation methods, while offering better inference-time efficiency.% and sample complexity.
    %\item Existing bias mitigation techniques show greater efficacy on \gls*{OccamNets} as compared to the traditional networks that violate Occam's principles.
\end{itemize}

\section{Related Work}

\textbf{Dataset Bias and Bias Mitigation.}  Deep networks trained with empirical risk minimization (ERM) tend to exploit training set biases resulting in poor test generalization~\cite{he2009learning,mehrabi2021survey,shrestha2022investigation,torralba2011unbiased}. Existing works for mitigating this problem have focused on these approaches: 1) focusing on rare data patterns through re-sampling~\cite{chawla2002smote,li2019repair}, 2) loss re-weighting~\cite{cui2019class,sagawa2020investigation}, 3)  adversarial debiasing~\cite{grand2019adversarial,kim2019learning}, 4) model ensembling~\cite{cadene2019rubi,clark2019don}, 5) minority/counterfactual sample generation~\cite{chawla2002smote,chen2020counterfactual,kim2021biaswap} and 6) invariant/robust risk minimization~\cite{arjovsky2019invariant,krueger2021out,sagawa2019distributionally}. Most of these methods require bias variables, e.g., sub-groups within a category, to be annotated~\cite{grand2019adversarial,kim2019learning,li2019repair,sagawa2020investigation,shrestha2022investigation}. Some recent methods have also attempted to detect and mitigate biases without these variables by training separate bias-amplified models for de-biasing the main model ~\cite{clark2020learning,nam2020learning,sanh2020learning,utama2020towards}. This paper is the first to explore architectural inductive biases for combating dataset bias. 

% \subsection{Dynamic Computations} 
% The goal of OccamNets is to dynamically choose input parts and network capacity to improve bias-resilience. Unfortunately, previous works on dynamic computations mainly focus on improving computational efficiency, and avoid studying bias-resistance~\cite{wang2018skipnet,chen2020dynamic,han2021dynamic}. Regardless, the following lines of research are relevant to our work.

\textbf{Early Exit Networks.} \gls*{OccamNet} is a multi-exit architecture designed to encourage later layers to focus on samples that earlier layers find difficult. Multi-exit networks have been studied in past work to speed up average inference time by minimizing the amount of compute needed for individual examples~\cite{chen2020learning,hu2020triple,teerapittayanon2016branchynet,wolczyk2021zero}, but their impact on bias-resilience has not been studied. In \cite{scardapane2020should}, a unified framework for studying early exit mechanisms was proposed, which included commonly used training paradigms~\cite{hettinger2017forward,lee2015deeply,szegedy2015going,venkataramani2015scalable} and biological plausibility~\cite{mostafa2018deep,nokland2016direct,nokland2019training}. During inference, multi-exit networks choose the earliest exit based on either a learned criterion~\cite{chen2020learning} or through a heuristic, e.g., exit if the confidence score is sufficiently high~\cite{duggal2020elf}, exit if there is low entropy~\cite{teerapittayanon2016branchynet}, or exit if there is agreement among multiple exits~\cite{zhou2020bert}. Recently, \cite{duggal2020elf} proposed early exit networks for long-tailed datasets; however, they used a class-balanced loss and did not study robustness to hidden covariates, whereas, \gls*{OccamNets} generalize to these hidden variables without oracle bias labels during training. 
% \kk{What is easy -- should we just say for some samples instead, or simple to keep the terminology consistant; Or maybe rephrase as: We hypothesize that some input samples require fewer complex interactive computations compared to others, and therefore allowing them to exit the computational graph of deep neural network early might avoid unnecessary computations, making the network less suceptible to overfitting to spurious correlations. Or something along thse lines}
 %Very few works~\cite{duggal2020elf} have studied multi-exit architectures in relation to bias-robustness \ck{if you mention this, you want to say only a single paper has looked at this and what's not sufficient about the paper, otherwise it doesn't look as novel here}.\rs{Qualified the notion of difficulty.}

\textbf{Exit Modules and Spatial Maps.} 
\gls*{OccamNets} are biased toward using fewer spatial locations for prediction, which we enable by using spatial activation maps~\cite{he2017mask,long2015fully,ronneberger2015u}. While most recent convolutional neural networks (CNNs) use global average pooling followed by a linear classification layer~\cite{he2016deep,howard2017mobilenets,huang2017densely}, alternative pooling methods have been proposed, including spatial attention~\cite{anderson2018bottom,guo2021attention,hu2018gather,xu2015show} and dynamic pooling~\cite{hu2018gather,jaderberg2015spatial,lee2016generalizing}. However, these methods have not been explored for their ability to combat bias mitigation, with existing bias mitigation methods adopting conventional architectures that use global average pooling instead. For \gls*{OccamNets}, each exit produces a class activation map, which is biased toward using fewer visual locations.

\section{OccamNets}
\begin{figure}[ht]
 \centering
 \footnotesize
    \includegraphics[width=0.9\linewidth]{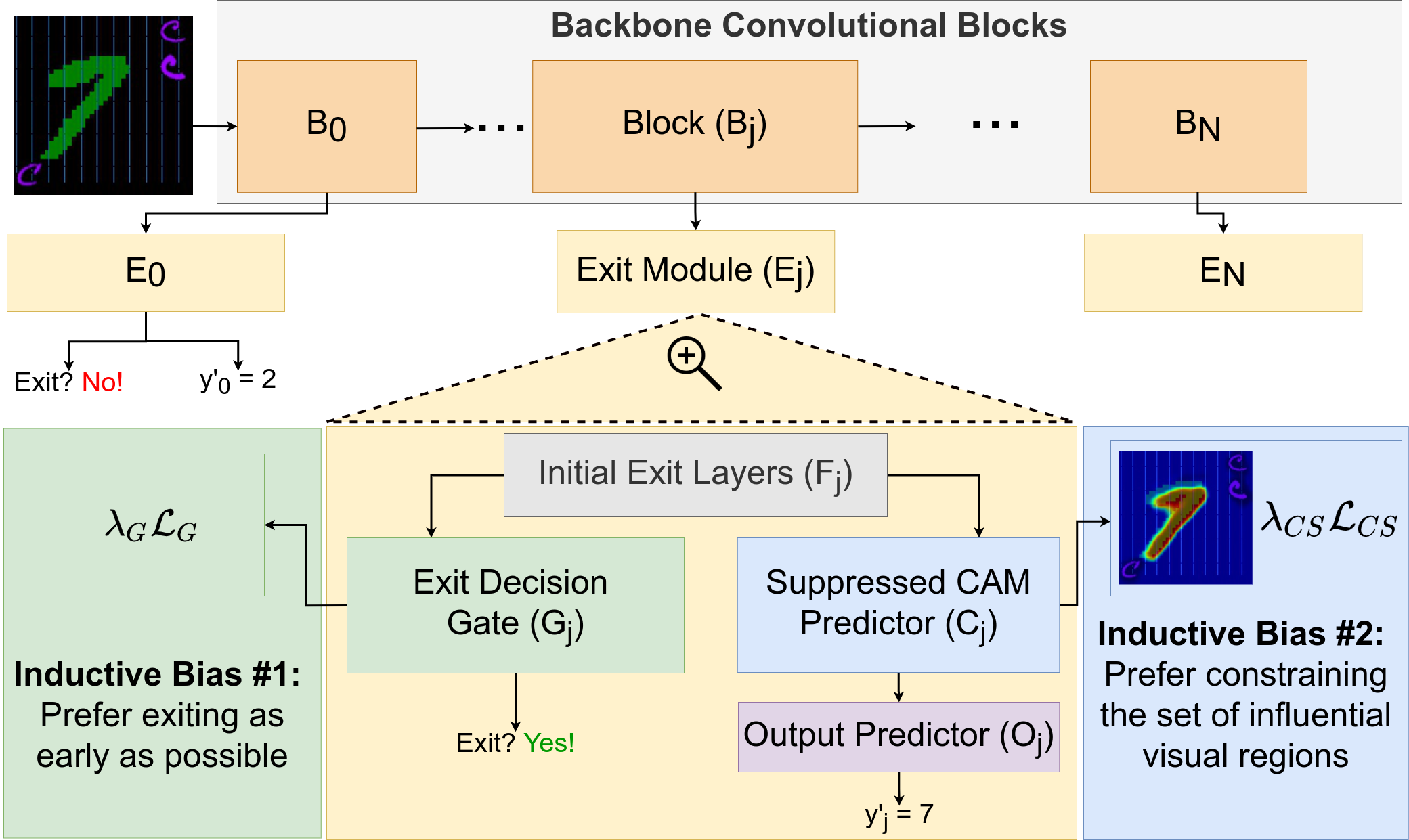}
    \caption{\gls*{OccamNets} are multi-exit architectures capable of exiting early through the exit decision gates. The exits yield class activation maps that are trained to use a constrained set of visual regions.}
     \label{fig:network}
\end{figure}

\subsection{OccamNet Architecture for Image Classification}
\gls*{OccamNets} have two inductive biases: a) they prefer exiting as early as possible, and b) they prefer using fewer visual regions for predictions. Following Occam's principles, we implement these inductive biases using simple, intuitive ideas: early exit is based on whether or not a sample is correctly predicted during training and visual constraint is based on suppressing regions that have low confidence towards the ground truth class. We implement these ideas in a CNN. Recent CNN architectures, such as ResNets~\cite{he2016deep} and DenseNets~\cite{huang2017densely}, consist of multiple blocks of convolutional layers. 
As shown in Fig.~\ref{fig:network}, these inductive biases are enabled by attaching an exit module $E_j$ to block $B_j$ of the CNN, as the blocks serve as natural endpoints for attaching them. Below, we describe how we implement these two inductive biases in \gls*{OccamNets}. %Later, we describe how the network is trained in  Secs.~\ref{sec:cam_suppression}, \ref{sec:task_loss} and \ref{sec:exit_decision_gates}.

%https://app.diagrams.net/#G1aBQp0ng-oMfmCQTVIOcns93x5u13b2Gx

In an \gls*{OccamNet}, each exit module $E_j$ takes in feature maps produced by the backbone network and processes them with $F_j$, which consists of two convolutional layers, producing feature maps used by the following components:

%Let us now introduce the components used in the exit modules $E_j$ of \gls*{OccamNets}. Each $E_j$ takes in feature maps yielded by the backbone network and processes them with an initial module with two convolutional layers $E_{init,j}$, which yields feature maps used by the following components:

%OccamNets are trained by combining all of the mentioned losses: \rs{Can fix the position at the end}

%The initial portion of each exit module consists of two $3 \times 3$ convolutional layers, followed by a $1 \times 1$ convolutional layer.

\textbf{Suppressed CAM Predictors ($C_j$).} Each $C_j$ consists of a single convolutional layer, taking in the feature maps from $F_{j}$ to yield class activation maps, $c$. The maps provide location wise predictions of classes. Following Occam's principles, the usage of visual regions in these CAMs is suppressed through the CAM suppression loss $\mathcal{L}_{CS}$ described in Sec.~\ref{sec:cam_suppression}.

\textbf{Output Predictors ($O_j$).} The output predictor applies global average pooling on the suppressed CAMs predicted by $C_j$ to obtain the output prediction vector, $\hat{y_j} \in \mathbb{R}^{n_Y}$, where $n_Y$ is the total number of classes. The entire network is trained with the output prediction loss~$\mathcal{L}_{O}$, which is a weighted sum of cross entropy losses between the ground truth $y$ and the predictions $\hat{y}_j$ from each of the exits. Specifically, the weighting scheme is formulated to encourage the deeper layers to focus on the samples that the shallower layers find difficult. The detailed training procedure is described in Sec.~\ref{sec:task_loss}. %For this, \gls*{OccamNets} use exit decision scores produced by exit decision gates (which are described below) to weigh the cross entropy losses. 

\textbf{Exit Decision Gates ($G_j$).}
During inference, OccamNet needs to decide whether or not to terminate the execution at $E_j$ on a per-sample basis. For this, each $E_j$ consists of an exit decision gate, $G_j$ that yields an exit decision score $g_j$, which is interpreted as the probability that the sample can exit from $E_j$. $G_j$ is realized via a ReLU layer followed by a sigmoid layer, taking in representations from $F_{j}$. The gates are trained via exit decision gate loss, $\mathcal{L}_G$ which is based on whether or not $O_j$ made correct predictions. The loss and the training procedure are elaborated further in Sec.~\ref{sec:exit_decision_gates}.

%Fig.~\ref{fig:losses} shows all of the loss terms used to train \gls*{OccamNets}.
The total loss used to train OccamNets is given by:
\begin{center}
    \includegraphics[width=0.9\linewidth]{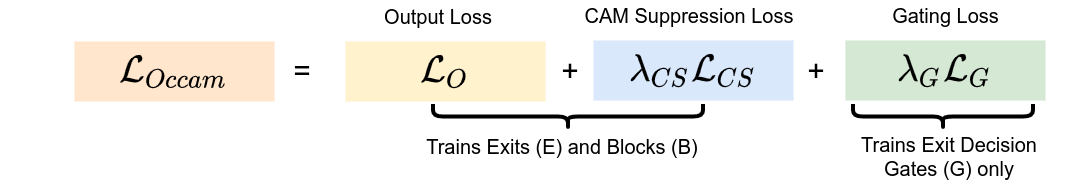}
\end{center}

\begin{comment}
\begin{figure}[h!]
 \centering
    \includegraphics[width=0.75\linewidth]{figures/losses.png}
      \caption{All the loss terms used to train \gls*{OccamNets}.}
      \label{fig:losses}
\end{figure}
\end{comment}

\subsection{Training the Suppressed CAMs}
\label{sec:cam_suppression}
To constrain the usage of visual regions, \gls*{OccamNets} regularize the CAMs so that only some of the cells exhibit confidence towards the ground truth class, whereas rest of the cells exhibit inconfidence i.e., have uniform prediction scores for all the classes. Specifically, let $c_y \in \mathbb{R}^{h \times w}$ be the CAM where each cell encodes the score for the ground truth class. Then, we apply regularization on the locations that obtain softmax scores lower than the average softmax score for the ground truth class. That is, let $\overline{c}_y$ be the softmax score averaged over all the cells in $c_y$, then the cells at location $l$, $c^l \in \mathbb{R}^{n_Y}$ are regularized if the softmax score for the ground truth class, $c^l_y$ is less than $\overline{c}_y$. The CAM suppression loss is:
\begin{align}
    \mathcal{L}_{CS} = \sum_{l=1}^{hw} \mathbbm{1} (c_y^l < \overline{c}_y) ~KLD(c^l, \frac{1}{n_Y}\mathbf{1}),
\end{align}
% \begin{align}
%     \mathcal{L}_{CS} = \sum_{l=1}^{hw} \mathbbm{1} (c_y^l < \tau) ~KLD(c^l, \frac{1}{n_Y}\mathbf{1}),
% \end{align}
where, $KLD(c^l, \frac{1}{n_Y}\mathbf{1})$ is the KL-divergence loss with respect to a uniform class distribution and $\mathbbm{1} (c_y^l < \overline{c}_y)$ ensures that the loss is applied only if the ground truth class scores lower than $\overline{c}_y$. The loss weight for $\mathcal{L}_{CS}$ is $\lambda_{CS}$, which is set to 0.1 for all the experiments.

\subsection{Training the Output Predictors}
\label{sec:task_loss}
%We train the exits with a loss weighting scheme that encourages the earliest exit to focus on bias-prone/easy samples, thereby, helping the later exits to focus on bias-conflicting/difficult samples. 
The prediction vectors $\hat{y}$ obtained by performing global average pooling on the suppressed CAMs $c$ are used to compute the output prediction losses. Specifically, we train a bias-amplified first exit $E_0$, using a loss weight of: ${W}_0 = p_0^{\gamma_0}$, where, $p_0$ is the softmax score for the ground truth class. Here, $\gamma_0 > 0$ encourages $E_0$ to amplify biases i.e., it provides higher loss weights for the samples that already have high scores for the ground truth class. This encourages $E_0$ to focus on the samples that it already finds easy to classify correctly. For all the experiments, we set $\gamma_0 = 3$ to sufficiently amplify the biases. The subsequent exits are then encouraged to focus on samples that the preceding exits find difficult. For this, the loss weights are defined as:
\begin{align}
    \mathcal{W}_j = (1 - g_{j-1} + \epsilon),~\text{if j  $>$ 0},
\end{align}
% \kk{This is not clear, clearly explain why scaling the probabalities by a positive factor means that they will be easier. ALso, what does loss weight mean?? Why is W0 not a part of eqn 5? Why is W0 not flowing through the netowrk M? Several things are unclear here}.
where, $g_{j-1}$ is the exit decision score predicted by $(j-1)^{th}$ exit decision gate and $\epsilon=0.1$ is a small offset to ensure that all the samples receive a minimal, non-zero loss weight. For the samples where $g_{j-1}$ is low, the weight loss for $j^{th}$ exit, $\mathcal{W}_j$ becomes high. The total output loss is then:
\begin{align}
    \mathcal{L}_O = \sum_{j=0}^{n_E - 1} \mathcal{W}_j ~ CE(\hat{y}_j, y),
\end{align}
where, $CE(\hat{y}_j, y)$ is the cross-entropy loss and $n_E$ is the total number of exits. Note that $E_j$'s are 0-indexed and the first bias-amplified exit $E_0$ is not used during inference. Furthermore, during training, we prevent the gradients of $E_0$ from passing through $B_0(.)$ to avoid degrading the representations available for the deeper blocks and exits.

\subsection{Training the Exit Decision Gates}
\label{sec:exit_decision_gates}
Each exit decision gate $G_j(.)$ yields an exit probability score $\hat{g_j} = G_j(.)$. During inference, samples with $\hat{g_j} \ge 0.5$ exit from $E_j$ and samples with $\hat{g}_j < 0.5$ continue to the next block $B_{j+1}$, if available. During training, all the samples use the entire network depth and $g_j$ is used to weigh losses as described in Sec.~\ref{sec:task_loss}. Now, we specify the exit decision gate loss used to train $G_j$:
\begin{align}
\mathcal{L}_G = \sum_{k \in \{0,1\}} \frac{\mathbbm{1}(g_j=k) BCE(g_j, \hat{g_j})}{\sqrt{\sum \mathbbm{1}(g_j=k)}},
\end{align}
where $g_j$ is the ground truth value for the $j^{th}$ gate, which is set to $1$ if the predicted class $y'$ is the same as the ground truth class $y$ and 0 otherwise. That is, $G_j$ is trained to exit if the sample is correctly predicted at depth $j$, else it is trained to continue onto the next block. Furthermore, the denominator: $\sqrt{\sum \mathbbm{1}(g_j=k)}$ balances out the contributions from the samples with $g=1$ and $g=0$ to avoid biasing one decision over the other. With this setup, sufficiently parameterized models that obtain $100\%$ training accuracy will result in a trivial solution where $g_j$ is always set to $1$ i.e., the exit will learn that all the samples can exit. To avoid this issue, we stop computing $g_j$ once $E_j$'s mean-per-class training accuracy reaches a predefined threshold $\tau_{acc,j}$. During training, we stop the gradients from $G$ from passing through $F_{j}(.)$ and $B(.)$, since this improved the training stability and overall accuracy in the preliminary experiments. The loss weight $\lambda_G$ is set to 1 in all the experiments.

\section{Experimental Setup}
%\kk{This para is probably not necessary} We describe the biased image classification datasets in Sec.~\ref{sec:datasets}, comparison methods in Sec.~\ref{sec:comparison-methods} and metrics, hyperparameters, and model selection criteria in Sec.~\ref{sec:metrics}.
 
\subsection{Datasets}
\label{sec:datasets}
\begin{figure}[ht]
\begin{subfigure}{.32\textwidth}
  \centering
  % include first image
  \includegraphics[width=0.95\linewidth]{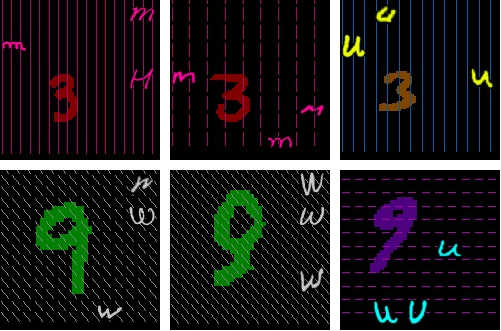}  
  \caption{\gls*{BiasedMNIST}}
  \label{fig:biased-mnist}
\end{subfigure}
\begin{subfigure}{.32\textwidth}
  \centering
  % include second image
  \includegraphics[width=0.95\linewidth]{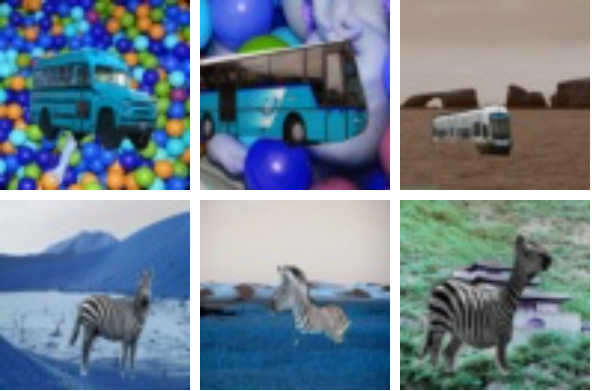}  
  \caption{\gls*{COCO}}
  \label{fig:coco}
\end{subfigure}
\begin{subfigure}{.32\textwidth}
  \centering
  % include second image
  \includegraphics[width=0.95\linewidth]{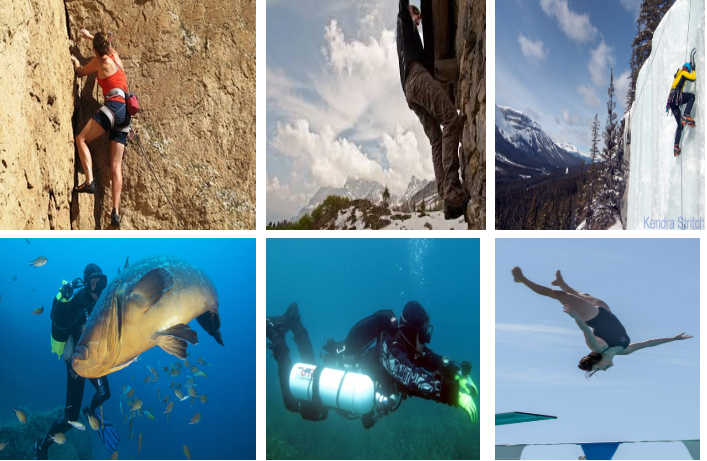}  
  \caption{\gls*{BAR}}
  \label{fig:BAR}
\end{subfigure}
\caption{ For each dataset, the first two columns show bias-aligned (majority) samples, and the last column shows bias-conflicting (minority) samples. For \gls*{BAR}, the train set does not contain any bias-conflicting samples.}
\label{fig:datasets}
\end{figure}

\textbf{\gls*{BiasedMNIST}~\cite{shrestha2022investigation}.} As shown in Fig.~\ref{fig:biased-mnist}, \gls*{BiasedMNIST} requires classifying MNIST digits while remaining robust to multiple sources of biases, including color, texture, scale, and contextual biases. This is more challenging than the widely used Colored MNIST dataset~\cite{arjovsky2019invariant,kim2019learning,li2019repair}, where the only source of bias is the spuriously correlated color. In our work, we build on the version created in \cite{shrestha2022investigation}. We use $160\times160$ images with $5\times5$ grids of cells, where the target digit is placed in one of the grid cells and is spuriously correlated with: a) digit size/scale (number of cells a digit occupies), b) digit color, c) type of background texture, d) background texture color, e) co-occurring letters, and f) colors of the co-occurring letters. Following~\cite{shrestha2022investigation}, we denote the probability with which each digit co-occurs with its biased property in the training set by $p_{bias}$. For instance, if $p_{bias}=0.95$, then 95\% of the digit 1s are red, 95\% of digit 1s co-occur with letter `a' (not necessarily colored red) and so on. We set $p_{bias}$ to 0.95 for all the experiments. The validation and test sets are unbiased. \gls*{BiasedMNIST} has 10 classes and 50K train, 10K validation, and 10K test samples.

\textbf{\gls*{COCO}~\cite{ahmed2020systematic}.} As shown in Fig.~\ref{fig:coco}, \gls*{COCO} puts COCO objects~\cite{lin2014microsoft} on spuriously correlated Places backgrounds~\cite{zhou2017places}. For instance, buses mostly appear in front of balloons and birds in front of trees. The dataset provides three different test sets: a) biased backgrounds (in-distribution), which reflects the object-background correlations present in the train set, b) unseen backgrounds (non-systematic shift), where the objects are placed on backgrounds that are absent from the train set and c) seen, but unbiased backgrounds (systematic shift) where the objects are placed on backgrounds that were not spuriously correlated with the objects in the train set.  Results in \cite{ahmed2020systematic} show it is difficult to maintain high accuracy on both the in-distribution and shifted-distribution test sets. Apart from that, \gls*{COCO} also includes an anomaly detection task, where anomalous samples from unseen object class need to be distinguished from the in-distribution samples. \gls*{COCO} has 9 classes with 7200 train, 900 validation, and 900 test images. 

\textbf{Biased Action Recognition (\gls*{BAR})~\cite{nam2020learning}.}
\gls*{BAR} reflects real world challenges where bias attributes are not explicitly labeled for debiasing algorithms, with the test set containing additional correlations not seen during training. The dataset consists of correlated action-background pairs, where the train set consists of selected action-background pairs, e.g., climbing on a rock, whereas the evaluation set consists of differently correlated action-background pairs, e.g., climbing on snowy slopes (see Fig.~\ref{fig:BAR}). The background is not labeled for the debiasing algorithms, making it a challenging benchmark. \gls*{BAR} has 6 classes with 1941 train and 654 test samples.

% \subsubsection{\gls*{CelebA}~\cite{liu2018large}} is a commonly used benchmark for testing bias robustness for image recognition models. It consists of celebrity faces annotated with different attributes. Following the conventional setup~\cite{sagawa2019distributionally,sagawa2020investigation,shrestha2022investigation}, we test the task of hair color prediction (blond or non-blond) where gender (male or non-male) acts as a bias. The challenge here is that most male celebrities have non-blond hair color, so methods fail on blond haired male celebrities.

\subsection{Comparison Methods, Architectures and Other Details}
\label{sec:comparison-methods}
We compare \gls*{OccamNets} with four state-of-the-art bias mitigation methods, apart from the vanilla empirical risk minimization procedure:

\begin{itemize}
    \item \textbf{Empirical Risk Minimization (\gls*{ERM})} is the default method used by most deep learning models and it often leads to dataset bias exploitation since it minimizes the train loss without any debiasing procedure.% \kk{Make it clear that is is the default system for most of deep networks...aka regular}
    
    \item \textbf{Spectral Decoupling (SD)~\cite{pezeshki2020gradient}} applies regularization to model outputs to help decouple features. This can help the model focus more on the signal.
    
    \item \textbf{Group Upweighting (\gls*{UpWt})} balances the loss contributions from the majority and the minority groups by multiplying the loss by $\frac{1}{n_g^\gamma}$, where $n_g$ is the number of samples in group $g$ and $\gamma$ is a hyper-parameter. 
    
    \item \textbf{Group DRO (\gls*{gDRO})~\cite{DBLP:journals/corr/abs-1911-08731}} is an instance of a broader family of distributionally robust optimization techniques~\cite{duchi2019distributionally,namkoong2016stochastic,rahimian2019distributionally}, that optimizes for the difficult groups in the dataset.% For \gls*{gDRO}, the goal is to optimize for the worst dataset group e.g., the rarest group of blond-haired male celebrities in \gls*{CelebA}.
    
    \item \textbf{Predictive Group Invariance (\gls*{PGI})~\cite{ahmed2020systematic}} is another grouping method, that encourages matched predictive distributions across easy and hard groups within each class. It penalizes the KL-divergence between predictive distributions from within-class groups.
\end{itemize}

\textbf{Dataset Sub-groups.} For debiasing, \gls*{UpWt}, \gls*{gDRO}, and \gls*{PGI} require additional labels for covariates (sub-group labels). Past work has focused on these labels being supplied by an oracle; however, having access to all relevant sub-group labels is often impractical for large datasets. Some recent efforts have attempted to infer these sub-groups. Just train twice (JTT)~\cite{liu2021just} uses a bias-prone ERM model by training for a few epochs to identify the difficult groups. Environment inference for invariant learning (EIIL)~\cite{creager2021environment} learns sub-group assignments that maximize the invariant risk minimization objective~\cite{arjovsky2019invariant}. Unfortunately, inferred sub-groups perform worse in general than when they are supplied by an oracle~\cite{ahmed2020systematic,liu2021just}. For the methods that require them, which \emph{excludes} \gls*{OccamNets}, we use oracle group labels (i.e., for \gls*{BiasedMNIST} and \gls*{COCO}). Inferred group labels are used for \gls*{BAR}, as oracle labels are not available.

%\kk{Mention discussion about how this is unrealistic and removing this previledged data access further weakens their performance.} \rs{Added something before saying we use it for strong comparison numbers.}

For \gls*{BiasedMNIST}, all the samples having the same class and the same value for all of the spurious factors are placed in a single group. For \gls*{COCO}, objects placed on spuriously correlated backgrounds form the majority group, while the rest form the minority group. \gls*{BAR} does not specify oracle group labels, so we adopt the JTT method. Specifically, we train an ERM model for single epoch, reserving 20\% of the samples with the highest losses as the difficult group and the rest as the easy group. We chose JTT over EIL for its simplicity. \gls*{OccamNets}, of course do not require such group labels to be specified.

\textbf{Architectures.} ResNet-18 is used as the standard baseline architecture for our studies. We compare it with an OccamNet version of ResNet-18, i.e., OccamResNet-18. To create this architecture, we add early exit modules to each of ResNet-18's convolutional blocks. To keep the number of parameters in OccamResNet-18 comparable to ResNet-18, we reduce the feature map width from 64 to 48. Assuming 1000 output classes, ResNet-18 has 12M parameters compared to 8M in OccamResNet-18. Further details are in the appendix.

%We add the proposed inductive biases to the standard ResNets, creating OccamResNets. We first run the comparison methods on the standard ResNet and compare them to OccamResNet. We also run the comparison methods on OccamResNet to see if the existing methods benefit from the proposed inductive biases. For the standard architecture, we choose ResNet18. To create OccamResNet18, we add the proposed exit modules at each of the four convolutional blocks. To keep the number of parameters in OccamResNet18 under that of ResNet, we reduce the feature map width from 64 to 48. Assuming 1000-class outputs, the total number of parameters are 11,689,512 and 7,939,860 for ResNet18 and OccamResNet18 respectively. For \gls*{COCO}, the images are small ($64 \times 64$), so following~\cite{ahmed2020systematic}, we replace the first convolutional  layer (kernel size=$7$, padding=$3$, stride=$2$), with a smaller layer (kernel size=$3$, padding=$1$ and stride=$1$) and also remove the initial max pooling layer. 

%\footnotetext{In \cite{ahmed2020systematic} Wide ResNet was used with a Tensorflow implementation. We could not replicate their results in PyTorch, so we used ResNet-18, which achieved comparable results on the systematic test set.}

\textbf{Metrics and Model Selection.}
\label{sec:metrics}
We report the means and standard deviations of test set accuracies computed across five different runs for all the datasets. For \gls*{BiasedMNIST}, we report the unbiased test set accuracy (i.e., $p_{bias} = 0.1$) alongside the majority and minority group accuracies for each bias variable. For \gls*{COCO}, unless otherwise specified, we report accuracy on the most challenging test split:  with seen, but unbiased backgrounds. We also report the average precision score to measure the ability to distinguish 100 anomalous samples from the in-distribution samples for the anomaly detection task of \gls*{COCO}. For \gls*{BAR}, we report the overall test accuracies. We use unbiased validation set of \gls*{BiasedMNIST} and validation set with unbiased backgrounds for \gls*{COCO} for hyperparameter tuning. The hyperparameter search grid and selected values are specified in the appendix.

%\textbf{Hyperparameters and Model Selection.} \gls*{BiasedMNIST} models are trained with Adam~\cite{kingma2014adam} for 90 epochs with a learning rate of $10^{-3}$ (decayed by 10 at epochs 50 and 70), batch size of 128 and weight decay of $5 \times 10^{-4}$. \gls*{COCO} models are trained with SGD for $150$ epochs at a learning rate of 0.1 (decayed by 10 at epochs: $100, 120, 140$), batch size of 64 and weight decay of $5 \times 10^{-4}$.  \gls*{BAR} models are trained with Adam for 150 epochs, with a learning rate of $5\times10^{-4}$, batch size of 64 and weight decay of $5\times10^{-4}$. For \gls*{OccamNets}, we set the exit-wise thresholds: $\tau_{acc}$ to \{0.5,0.6,0.7,0.8\} for \gls*{BiasedMNIST} and \gls*{COCO} and \{0.1,0.2,0.3,0.4\} for \gls*{BAR}. We tune the hyperparameters for \gls*{BiasedMNIST} using the unbiased validation set and for \gls*{COCO} using the systematically shifted validation set. \gls*{BAR} does not provide a validation set. We use early stopping for all the methods. Hyperparameter search grids and the selected values are provided in the appendix.  

%\kk{Section 4 is fine; There are places it could be improved and there is room for a lot of compression if needed, but nothing major like Section 3, so I left it mostly alone for now.}

\section{Results and Analysis}
%We compare the proposed \gls*{OccamNets} to the existing state-of-the-art bias mitigation methods. We also examine the compatibility between \gls*{OccamNets} and the comparison methods. Furthermore, we analyze the impacts of each of the proposed inductive biases. We end the section by studying robustness to different types of shifts and accuracy on datasets with less bias.

\subsection{Overall Results}
\label{sec:results}
\begin{table}[t]
\centering
\footnotesize
\caption{Unbiased test set accuracies comparing \gls*{OccamResNet} to the more conventional \gls*{ResNet} architectures without early exits and constrained class activation maps. We format the $\textbf{\underline{first}}$, $\textbf{second}$ and $\underline{third}$ best results. }
\label{tab:overall}
\addtolength\tabcolsep{5pt}
\begin{tabular}{lccc}
 \hline
Architecture+Method &
  \begin{tabular}[c]{@{}l@{}}\gls*{BiasedMNIST}\end{tabular} &
  \begin{tabular}[c]{@{}l@{}}\gls*{COCO}\end{tabular} &
  \begin{tabular}[c]{@{}l@{}}\gls*{BAR} \end{tabular} \\ \hline
\multicolumn{4}{c}{\textit{Results on Standard \gls*{ResNet}-18}}                                                             \\
ResNet+ERM  & 36.8 \tiny{$\pm$ 0.7} & \underline{35.6} \tiny{$\pm$ 1.0}  & \underline{51.3} \tiny{$\pm$1.9}    \\
ResNet+\gls*{SD}~\cite{pezeshki2020gradient} & 37.1 \tiny{$\pm$ 1.0} & 35.4 \tiny{$\pm$ 0.5}  & \underline{51.3} \tiny{$\pm$2.3}    \\
ResNet+\gls*{UpWt} & \underline{37.7} \tiny{$\pm$ 1.6} & 35.2 \tiny{$\pm$ 0.4}  & 51.1 \tiny{$\pm$1.9}   \\
ResNet+\gls*{gDRO}~\cite{sagawa2019distributionally} & 19.2 \tiny{$\pm$ 0.9} & 35.3 \tiny{$\pm 0.1$}  & 38.7 \tiny{$\pm$2.2}  \\
ResNet+\gls*{PGI}~\cite{ahmed2020systematic} & \textbf{48.6} \tiny{$\pm$ 0.7} & \textbf{42.7} \tiny{$\pm$ 0.6} & \textbf{\underline{53.6}} \tiny{$\pm$0.9} \\ 
\hline
\multicolumn{4}{c}{\textit{Results on \gls*{OccamResNet}-18}}                                                             \\
\gls*{OccamResNet} & \textbf{\underline{65.0}} \tiny{$\pm 1.0$} & \textbf{\underline{43.4}} \tiny{$\pm$ 1.0} & \textbf{52.6} \tiny{$\pm$1.9} \\
%\gls*{OccamResNet}+PGI & \textbf{\underline{69.6}} \tiny{$\pm 0.7$} & \textbf{\underline{43.6}} \tiny{$\pm$ 0.6} & \textbf{\underline{55.9}} \tiny{$\pm$0.7} \\
\hline
\end{tabular}%
\end{table}
% If there is some combination (ours+XXX) does really good, We could separate Ours into two lines, Ours: Occament and Ours:Final (Occamnet+PGI) perhaps...

\textbf{OccamNets vs. ERM and Recent Bias Mitigation Methods.} 
To examine how \gls*{OccamNets} fare against ERM and state-of-the-art bias mitigation methods, we run the comparison methods on ResNet and compare the results with \gls*{OccamResNet}. Results are given in  Table~\ref{tab:overall}. \gls*{OccamResNet} outperforms state-of-the-art methods on \gls*{BiasedMNIST} and \gls*{COCO} and rivals \gls*{PGI} on \gls*{BAR}, demonstrating that architectural inductive biases alone can help mitigate dataset bias. The gap between \gls*{OccamResNet} and other methods is large on \gls*{BiasedMNIST} (16.4 - 46.0\% absolute difference). For \gls*{COCO}, \gls*{PGI} rivals \gls*{OccamResNet}, and clearly outperforms all other methods, in terms of accuracy on the test split with seen, but unbiased backgrounds. \gls*{OccamResNet}'s results are impressive considering that Up Wt, gDRO, and PGI all had access to the bias group variables, unlike OccamNet, ERM, and SD.

%Furthermore, as shown in the last row, running \gls*{PGI} on \gls*{OccamResNet} further improves accuracy on all of the datasets. \ck{I am not sure we should include all of these results in the table, I think we could just in text say that while PGI performs slightly better than it, as we show in the next section combining them results in superior performance}

\textbf{Combining OccamNets with Recent Bias Mitigation Methods.} 
Because \gls*{OccamNets} are a new network architecture, we used OccamResNet-18 with each of the baseline methods instead of ResNet-18. These results are shown in Table~\ref{tab:occam-gains}, where we provide unbiased accuracy along with any \textcolor{gain}{improvement} or \textcolor{loss}{impairment} of performance when OccamResNet-18 is used instead of ResNet-18. All methods benefit from using the \gls*{OccamResNet} architecture compared to ResNet-18, with gains of 10.6\% - 28.2\% for \gls*{BiasedMNIST}, 0.9\% - 7.8\% for \gls*{COCO}, and 1.0\% - 14.2\% for \gls*{BAR}.

\begin{table}[t]
% \usepackage{graphicx}
%\begin{table}[]
\centering
\footnotesize
\caption{Unbiased accuracies alongside ~\textcolor{gain}{improvement}/~\textcolor{loss}{impairment} when the comparison methods are run on \gls*{OccamResNet} instead of \gls*{ResNet}.}
%\kk{This is unbiased acc.right? Need to specify}
\label{tab:occam-gains}
\addtolength\tabcolsep{3pt}
% \resizebox{\textwidth}{!}{%
\begin{tabular}{lccl}
\hline
Architecture+Method
 &
  \multicolumn{1}{c}{\begin{tabular}[c]{@{}c@{}}\gls*{BiasedMNIST}\end{tabular}} &
  \multicolumn{1}{c}{\begin{tabular}[c]{@{}c@{}}\gls*{COCO}\end{tabular}} &
  \multicolumn{1}{c}{\gls*{BAR}} \\ \hline
\gls*{OccamResNet} & \underline{65.0}~\textcolor{gain}{(+28.2)} & \textbf{43.4}~\textcolor{gain}{(+7.8)} & \underline{52.6}~\textcolor{gain}{(+1.3)} \\
\gls*{OccamResNet}+\gls*{SD}~\cite{pezeshki2020gradient}   & 55.2~\textcolor{gain}{(+18.1)} & 39.4~\textcolor{gain}{(+4.0)} & 52.3~\textcolor{gain}{(+1.0)} \\
\gls*{OccamResNet}+\gls*{UpWt} & \textbf{65.7}~\textcolor{gain}{(+28.0)} & \underline{42.9}~\textcolor{gain}{(+7.7)} & 52.2~\textcolor{gain}{(+1.1)} \\
\gls{OccamResNet}+\gls*{gDRO}~\cite{sagawa2019distributionally} & 29.8~\textcolor{gain}{(+10.6)} & 40.7~\textcolor{gain}{(+5.4)} & \textbf{52.9}~\textcolor{gain}{(+14.2)} \\
\gls*{OccamResNet}+\gls*{PGI}~\cite{ahmed2020systematic} & \textbf{\underline{69.6}}~\textcolor{gain}{(+21.0)} & \textbf{\underline{43.6}}~\textcolor{gain}{(+0.9)} & \textbf{\underline{55.9}}~\textcolor{gain}{(+2.3)} \\ \hline
\end{tabular}%
% }
%\end{table}
\end{table}

\subsection{Analysis of the Proposed Inductive Biases}

\begin{figure}[ht!]
\begin{subfigure}{.32\textwidth}
  \centering
  % include first image
  \includegraphics[width=0.98\linewidth]{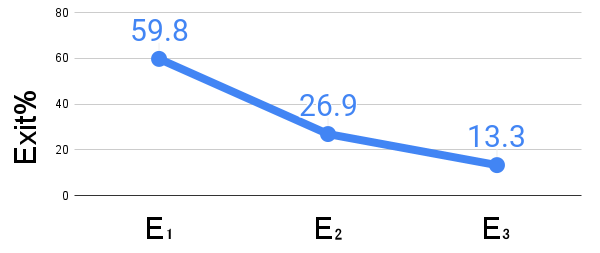}  
  \caption{\gls*{BiasedMNIST}}
  \label{fig:biased-mnist-exits}
\end{subfigure}
\begin{subfigure}{.32\textwidth}
  \centering
  % include second image
  \includegraphics[width=0.98\linewidth]{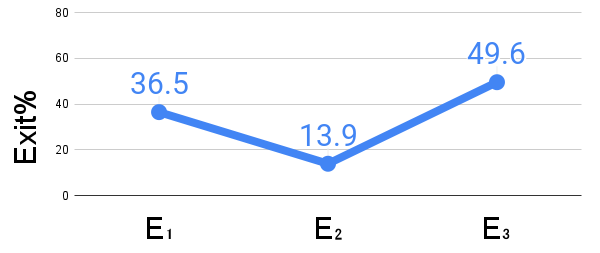}  
  \caption{\gls*{COCO}}
  \label{fig:coco-exits}
\end{subfigure}
\begin{subfigure}{.32\textwidth}
  \centering
  % include second image
  \includegraphics[width=0.98\linewidth]{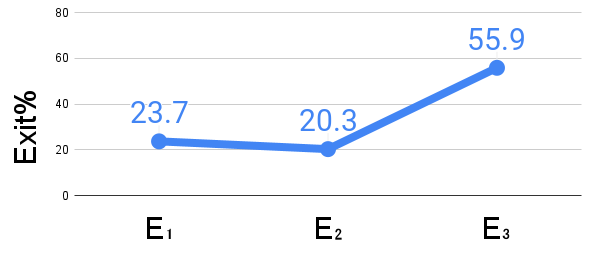}  
  \caption{\gls*{BAR}}
  \label{fig:BAR-exits}
\end{subfigure}
\caption{Percentage of samples exited (Exit\%) from each exit (barring $E_0$).}
\label{fig:exit-stats}
\end{figure}

In this section, we analyze the impacts of each of the proposed modifications in \gls*{OccamNets} and their success in achieving the desired behavior.

\textbf{Analysis of Early Exits.}  \gls*{OccamResNet} has four exits, the first exit is used for bias amplification and the rest are used to potentially exit early during inference. To analyze the usage of the earlier exits, we plot the percentage of samples that exited from each exit in Fig.~\ref{fig:exit-stats}. For \gls*{BiasedMNIST} dataset, a large portion of the samples, i.e., 59.8\% exit from the shallowest exit of $E_1$ and only 13.3\% exit from the final exit $E_3$. For \gls*{COCO} and \gls*{BAR}, 50.4\% and 44.1\% samples exit before $E_3$, with 49.6\% and 55.9\% samples using the full depth respectively. These results show that the \gls*{OccamNet}s favor exiting early, but that they do use the full network depth if necessary.

\begin{figure}[h]
 \centering
    \includegraphics[width=0.99\linewidth]{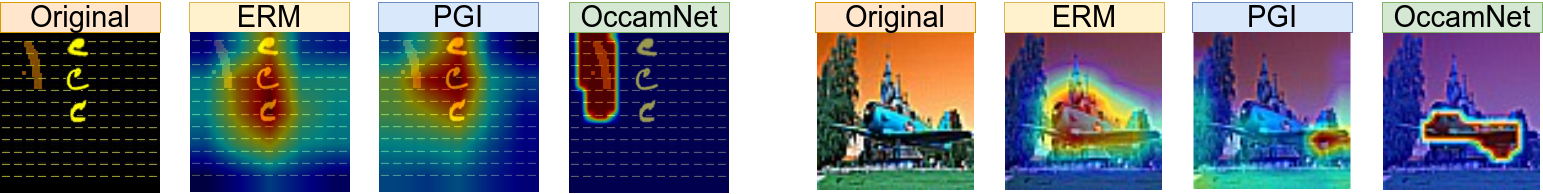}
      \caption{Original image, and Grad-CAM visualizations for ERM and PGI on ResNet, and CAM visualizations on OccamResNet. The visualizations are for the ground truth.}
      \label{fig:cams}
\end{figure}
\textbf{CAM Visualizations.}  To compare the localization capabilities of OccamResNets to ResNets, we present CAM visualizations in Fig.~\ref{fig:cams}. For ResNets that were run with ERM and PGI, we show Grad-CAM visualizations~\cite{selvaraju2017grad}, whereas for OccamResNets, we directly visualize the CAM heatmaps obtained from the earliest exit used for each sample. As shown in the figure, OccamResNet generally prefers smaller regions that include the target object. On the other hand, comparison methods tend to focus on larger visual regions that include irrelevant object/background cues leading to lower accuracies.
%\kk{There is no analysis here, maybe just say visualization of ...}
% We compare class activation maps between \gls*{ResNet}, \gls*{OccamResNet} with and without pooling losses in Fig. XXX. Clearly, $\mathcal{L}_{CAM}$ helps the model focus on fewer visual regions. We also hypothesize that it acts as a regularizer, to regularize model confidence as in~\cite{pezeshki2020gradient}. As shown by the ablation study in Table~\ref{}, model does worse without pooling loss.

% \begin{figure}[ht]
% \begin{subfigure}{1.0\textwidth}
%   \centering
%   \includegraphics[width=0.45\linewidth]{figures/bmnist-cam-good-7.jpg}
%   \hspace{10pt}
%   \includegraphics[width=0.45\linewidth]{figures/bmnist-cam-good-9.jpg}  
%   \caption{\gls*{BiasedMNIST}}
%   \label{fig:bmnist-cam-good}
% \end{subfigure}
% \begin{subfigure}{1.0\textwidth}
%   \centering
%   \includegraphics[width=0.45\linewidth]{figures/coco-cam-good-dog.jpg}
%   \hspace{10pt}
%   \includegraphics[width=0.45\linewidth]{figures/coco-cam-good-bus.jpg}  
%   \caption{\gls*{COCO}}
%   \label{fig:coco-cam-good}
% \end{subfigure}
% \caption{Original image, and grad cam visualizations for ERM and PGI on ResNet, and CAM visualization for OccamTrainer.}
% \label{fig:cams}
% \end{figure}

\begin{table}[]
\centering
\caption{Ablation Studies on \gls*{OccamResNet}}
\label{tab:ablations}
\addtolength\tabcolsep{1.5pt}
\begin{tabular}{lccc}
 \hline
Ablation Description &
  \begin{tabular}[c]{@{}l@{}}\gls*{BiasedMNIST}\end{tabular} &
  \begin{tabular}[c]{@{}l@{}}\gls*{COCO}\end{tabular}  \\ \hline
Using only one exit at the end & 35.9 & 35.0 \\
Weighing all the samples equally i.e., $\mathcal{W}_j = 1$  & \textbf{\underline{66.3}} & \textbf{40.8} \\
Without the CAM suppression loss i.e., $\lambda_{CS} = 0$ & \underline{48.2} & \underline{37.1} \\
\hline
Full \gls*{OccamResNet} & \textbf{65.0} & \textbf{\underline{43.4}} \\\hline
\end{tabular}%
\end{table}

%\kk{Please use consistant terminology, I don't know what gate-weighted loss here is...I know this is incomplete, but just wanted to give some early feedback.}
\textbf{Ablations.} To study the importance of the proposed inductive biases, we perform ablations on \gls*{BiasedMNIST} and \gls*{COCO}. First, to examine if the multi-exit setup is helpful, we train networks with single exit attached to the end of the network. This caused accuracy drops of 29.1\% on \gls*{BiasedMNIST} and 8.4\% on \gls*{COCO}, indicating that the multi-exit setup is critical. To examine if the weighted output prediction losses are helpful or harmful, we set all the loss weights ($\mathcal{W}_j$) to 1. This resulted in an accuracy drop of 2.6\% on \gls*{COCO}. However, it improved accuracy on \gls*{BiasedMNIST} by 1.3\%. We hypothesize that since the earlier exits suffice for a large number of samples in \gls*{BiasedMNIST} (as indicated in Fig.~\ref{fig:biased-mnist-exits}), the later exits may not receive sufficient training signal with the weighted output prediction losses. Finally, we ran experiments without the CAM suppression loss by setting $\lambda_{CS}$ to 0. This caused large accuracy drops of 16.8\% on \gls*{BiasedMNIST} and 6.3\% on \gls*{COCO}. These results show that both inductive biases are vital for \gls*{OccamNets}.

\subsection{Robustness to Different Types of Shifts} 
A robust system must handle different types of bias shifts. To test this ability, we examine the robustness to each  bias variable in \gls*{BiasedMNIST} and we also compare the methods on the differently shifted test splits of \gls*{COCO}.
% \usepackage{multirow}
% \usepackage{graphicx}
\begin{table*}[]
\centering
\caption{Accuracies on majority (maj)/minority (min) groups for each bias variable in \gls*{BiasedMNIST} ($p_{bias}=0.95$). We \textbf{embolden} the results with the lowest differences between the groups.}
\label{tab:biased-mnist}
\addtolength\tabcolsep{1.75pt}
\resizebox{\textwidth}{!}{%
\begin{tabular}{lccccccccccccc} \hline
\multicolumn{1}{l}{\multirow{2}{*}{\begin{tabular}[c]{@{}l@{}}Architecture+Method\end{tabular}}} &
  \multicolumn{1}{l}{} &
  \multicolumn{2}{c}{\begin{tabular}[c]{@{}c@{}}Digit \\ Scale\end{tabular}} &
  \multicolumn{2}{c}{\begin{tabular}[c]{@{}c@{}}Digit\\ Color\end{tabular}} &
  \multicolumn{2}{c}{\begin{tabular}[c]{@{}c@{}}Texture\end{tabular}} &
  \multicolumn{2}{c}{\begin{tabular}[c]{@{}c@{}}Texture \\Color\end{tabular}} &
  \multicolumn{2}{c}{\begin{tabular}[c]{@{}c@{}}Letter\end{tabular}} &
  \multicolumn{2}{c}{\begin{tabular}[c]{@{}c@{}}Letter \\ Color\end{tabular}} \\
\multicolumn{1}{l}{} &
  \multicolumn{1}{l}{\begin{tabular}[c]{@{}l@{}}Test \\ Acc.\end{tabular}} &
  \multicolumn{2}{c}{maj/min} &
  \multicolumn{2}{c}{maj/min} &
  \multicolumn{2}{c}{maj/min} &
  \multicolumn{2}{c}{maj/min} &
  \multicolumn{2}{c}{maj/min} &
  \multicolumn{2}{c}{maj/min}
  \\ \hline
 \multicolumn{1}{l}{ResNet+ERM} &
  \multicolumn{1}{l}{36.8} &
  \multicolumn{2}{c}{87.2/31.3} &
  \multicolumn{2}{c}{78.5/32.1} &
  \multicolumn{2}{c}{76.1/32.4} &
  \multicolumn{2}{c}{41.9/36.3} &
  \multicolumn{2}{c}{46.7/35.7} &
  \multicolumn{2}{c}{45.7/35.9} \\
  \multicolumn{1}{l}{ResNet+PGI} &
  \multicolumn{1}{l}{48.6} &
  \multicolumn{2}{c}{91.9/43.8} &
  \multicolumn{2}{c}{84.8/44.6} &
  \multicolumn{2}{c}{79.5/45.1} &
  \multicolumn{2}{c}{51.3/48.3} &
  \multicolumn{2}{c}{67.2/46.5} &
  \multicolumn{2}{c}{55.8/47.9} \\\hline
  \multicolumn{1}{l}{OccamResNet} &
  \multicolumn{1}{l}{65.0} &
  \multicolumn{2}{c}{94.6/61.7} &
  \multicolumn{2}{c}{96.3/61.5} &
  \multicolumn{2}{c}{81.6/63.1} &
  \multicolumn{2}{c}{\textbf{66.8/64.8}} &
  \multicolumn{2}{c}{\textbf{64.7/65.1}} &
  \multicolumn{2}{c}{\textbf{64.7/65.1}} \\
  \multicolumn{1}{l}{OccamResNet+PGI} &
  \multicolumn{1}{l}{69.6} &
  \multicolumn{2}{c}{95.4/66.7} &
  \multicolumn{2}{c}{97.0/66.5} &
  \multicolumn{2}{c}{88.6/67.4} &
  \multicolumn{2}{c}{\textbf{71.4/69.4}} &
  \multicolumn{2}{c}{\textbf{69.6/69.6}} &
  \multicolumn{2}{c}{\textbf{70.5/69.5}} \\\hline

\end{tabular}}
\end{table*}
% \usepackage{graphicx}
\begin{table}[]
\centering
\caption{Accuracies on all three test splits of \gls*{COCO}, alongside mean average precision for the anomaly detection task.} %\kk{Are the numbers up to date; COCO on places for PGI+ours is supposed to only give +0.9. Which numbers does that correspond to? What is reported in Table 2 and 3?} \rs{I was reporting systematic shift acc on previous tables, since that is the most challenging one. I should make it clearer.}}
\label{tab:coco-on-places}
\addtolength\tabcolsep{2pt}
\begin{tabular}{lcccc}
 \hline
Architecture+Method &
  \begin{tabular}[c]{@{}c@{}}Biased\\ Backgrounds\end{tabular} &
  \begin{tabular}[c]{@{}c@{}}Unseen \\ Backgrounds\end{tabular} &
  \begin{tabular}[c]{@{}c@{}}Seen, but \\ Non-Spurious \\ Backgrounds\end{tabular} &
  \begin{tabular}[c]{@{}c@{}}Anomaly \\ Detection\end{tabular} \\ \hline
\multicolumn{5}{c}{\textit{Results on Standard ResNet-18}}                                                             \\
ResNet+ERM  & \textbf{\underline{84.9}} \tiny{$\pm$ 0.5} & \underline{53.2} \tiny{$\pm$0.7} & 35.6 \tiny{$\pm$1.0} & 20.1 \tiny{$\pm$1.5} \\
ResNet+PGI~\cite{ahmed2020systematic} & 77.5 \tiny{$\pm$0.6} & 52.8 \tiny{$\pm$0.7} & \underline{42.7} \tiny{$\pm$0.6} &  \underline{20.6} \tiny{$\pm$2.1} \\ 
\hline
\multicolumn{5}{c}{\textit{Results on \gls*{OccamResNet}-18}}                             \\
\gls*{OccamResNet} & \textbf{84.0} \tiny{$\pm$1.0}  & \textbf{\underline{55.8}} \tiny{$\pm$1.2} & \textbf{43.4} \tiny{$\pm$1.0} &  \textbf{\underline{22.3}} \tiny{$\pm$2.8}  \\
\gls*{OccamResNet}+\gls*{PGI}~\cite{ahmed2020systematic} & \underline{82.8} \tiny{$\pm$0.6} & \textbf{55.3} \tiny{$\pm$1.3} & \textbf{\underline{43.6}} \tiny{$\pm$ 0.6}	&  \textbf{21.6} \tiny{$\pm$1.6} \\\hline
\end{tabular}%
\end{table}
% \kk{The first column needs to show second and third place,}

In Table~\ref{tab:biased-mnist}, we compare how \gls*{ResNet} and \gls*{OccamResNet} are affected by the different bias variables in \gls*{BiasedMNIST}. For this, we present majority and minority group accuracies for each variable. Bias variables with large differences between the majority and the minority groups, i.e., large majority/minority group discrepancy (MMD) are the most challenging spurious factors.  \gls*{OccamResNet}s, with and without PGI, improve on both majority and minority group accuracies across all the bias variables. OccamResNets are especially good at ignoring the distracting letters and their colors, obtaining MMD values between 0-1\%. ResNets, on the other hand are susceptible to those spurious factors, obtaining MMD values between 7.9-20.7\%. Among all the variables, digit scale and digit color are the most challenging ones and OccamResNets mitigate their exploitation to some extent. Next, we show accuracies for base method and PGI on both ResNet-18 and OccamResNet-18 on all of the test splits of \gls*{COCO} in Table~\ref{tab:coco-on-places}. The different test splits have different kinds of object-background combinations, and ideally the method should work well on all three test splits. \gls*{PGI} run on ResNet-18 improves on the split with seen, but non-spurious backgrounds but incurs a large accuracy drop of 7.4\% on the in-distribution test set, with biased backgrounds. On the other hand, OccamResNet-18 shows only 0.9\% drop on the biased backgrounds, while showing 2.6\% accuracy gains on the split with unseen backgrounds and 7.8\% accuracy gains on the split with seen, but non-spurious backgrounds. It further obtains 2.2\% gains on the average precision metric for the anomaly detection task. PGI run on OccamResNet exhibits a lower drop of 2.1\% on the in-distribution split as compared to the PGI run on ResNet, while obtaining larger gains on rest of the splits. These results exhibit that \gls*{OccamNets} obtain high in-distribution and shifted-distribution accuracies.

\subsection{Evaluation on Other Architectures}
To examine if the proposed inductive biases improve bias-resilience in other architectures too, we created OccamEfficientNet-B2 and OccamMobileNet-v3 by modifying EfficientNet-B2~\cite{tan2019efficientnet} and MobileNet-v3~\cite{howard2019searching,howard2017mobilenets}. \gls*{OccamNet} variants outperform standard architectures on both \gls*{BiasedMNIST} (OccamEfficientNet-B2: 59.2 vs. EfficientNet-B2: 34.4 and OccamMobileNet-v3: 49.9 vs. MobileNet-v3: 40.4) and \gls*{COCO} (OccamEfficientNet-B2: 39.2 vs. EfficientNet-B2: 34.2 and OccamMobileNet-v3: 40.1 vs. MobileNet-v3: 34.9). The gains show that the proposed modifications help other architectures too.% We provide the detailed results in Sec.~\ref{sec:other_archs}.

\subsection{Do OccamNets Work on Less Biased Datasets?}

To examine if \gls*{OccamNets} also work well on datasets with less bias, we train ResNet-18 and OccamResNet-18 on 100 classes of the ImageNet dataset~\cite{deng2009imagenet}. OccamResNet-18 obtains competitive numbers compared to the standard ResNet-18 (OccamResNet-18: 92.1, vs. ResNet-18: 92.6, top-5 accuracies). However, as described in rest of this paper, OccamResNet-18 achieves this with improved resistance to bias (e.g., if the test distributions were to change in future). Additionally, it also reduces the computations with 47.6\% of the samples exiting from $E_1$ and 13.3\% of samples exiting from $E_2$. As such,~\gls*{OccamNets} have the potential to be the \textit{de facto} network choice for visual recognition tasks regardless of the degree of bias.

\section{Discussion}
%\subsection{Remarks on Model Selection}
%For preliminary experiments, we tuned the hyperparameters for COCO-on-PLACES. However, did not need to tune them further on other datasets. The hyperparameters used by full \gls*{OccamNets} include: training accuracy threshold: where intuitve values of $60\% to 80\%$ work well across datasets, and loss weight for CAM which is always set to $0.1$. On the other hand, some of the comparison methods were highly sensitive. For instance, $PGI$ had to be tuned on a range $0.1, 1, 10, 100$, and different datasets required different hyperparameters. \gls*{gDRO} was especially difficult to tune despite tuning learning rate, weight decay, balance factor, step size and the oracle group labels.  

\textbf{Relation to Mixed Capacity Models.} Recent studies show that sufficiently simple models, e.g., with fewer parameters~\cite{clark2020learning,hooker2020characterising} or models trained for a few epochs~\cite{liu2021just,utama2020towards} amplify biases. Specifically, \cite{hooker2020characterising} shows that model compression disproportionately hampers minority samples. Seemingly, this is an argument against smaller models (simpler hypotheses), i.e., against Occam's principles. However, \cite{hooker2020characterising} does not study network depth, unlike our work. Our paper suggests that using only the necessary capacity for each example yields greater robustness. 

%\textbf{Advantages of CAMs.}
%Free visualizations.
 %One could extend this to be supervised e.g., if the ground truth object localization is available, however we do not explore this in our work.

\textbf{Relation to Multi-Hypothesis Models.} Some recent works generate multiple plausible hypotheses~\cite{lee2022diversify,teney2021evading} and use extra information at test time to choose the best hypothesis. The techniques include training a set of models with dissimilar input gradients~\cite{teney2021evading} and training multiple prediction heads that disagree on a target distribution~\cite{lee2022diversify}. An interesting extension of \gls*{OccamNets} could be making diverse predictions through the multiple exits and through CAMs that focus on different visual regions. This could help avoid discarding complex features in favor of simpler ones~\cite{shah2020pitfalls}. This may also help with under-specified tasks, where there are equally viable ways of making the predictions~\cite{lee2022diversify}.

\textbf{Other Architectures and Tasks.}
We tested \gls*{OccamNets} implemented with CNNs; however, they may be beneficial to other architectures as well. The ability to exit dynamically could be used with transformers, graph neural networks, and feed-forward networks more generally. There is some evidence already for this on natural language inference tasks, where early exits improved robustness in a transformer architecture~\cite{zhou2020bert}. While the spatial bias is more vision specific, it could be readily integrated into recent non-CNN approaches for image classification~\cite{dosovitskiy2020image,liu2021swin,tolstikhin2021mlp,yu2021metaformer}.

\textbf{Conclusion.}  In summary, the proposed \gls*{OccamNets} have architectural inductive biases favoring simpler solutions. The experiments show improvements over state-of-the-art bias mitigation techniques. Furthermore, existing methods tend to do better with \gls*{OccamNets} as compared to the standard architectures.

% -------------------------------------------------
\subsubsection*{Acknowledgments.}
This work was supported in part by NSF awards \#1909696 and \#2047556. The views and conclusions contained herein are those of the authors and should not be interpreted as representing the official policies or endorsements of any sponsor. 

%\clearpage
\bibliographystyle{splncs04}
\bibliography{egbib}

\begin{thebibliography}{10}
\providecommand{\url}[1]{\texttt{#1}}
\providecommand{\urlprefix}{URL }
\providecommand{\doi}[1]{https://doi.org/#1}

\bibitem{adeli2019bias}
Adeli, E., Zhao, Q., Pfefferbaum, A., Sullivan, E., Fei-Fei, L., Niebles, J.C., Pohl, K.: Bias-resilient neural network. ArXiv  \textbf{abs/1910.03676} (2019)

\bibitem{agrawal2018don}
Agrawal, A., Batra, D., Parikh, D., Kembhavi, A.: Don't just assume; look and answer: Overcoming priors for visual question answering. In: 2018 {IEEE} Conference on Computer Vision and Pattern Recognition, {CVPR} 2018, Salt Lake City, UT, USA, June 18-22, 2018. pp. 4971--4980. {IEEE} Computer Society (2018). \doi{10.1109/CVPR.2018.00522}

\bibitem{ahmed2020systematic}
Ahmed, F., Bengio, Y., van Seijen, H., Courville, A.: Systematic generalisation with group invariant predictions. In: International Conference on Learning Representations (2020)

\bibitem{anderson2018bottom}
Anderson, P., He, X., Buehler, C., Teney, D., Johnson, M., Gould, S., Zhang, L.: Bottom-up and top-down attention for image captioning and visual question answering. In: 2018 {IEEE} Conference on Computer Vision and Pattern Recognition, {CVPR} 2018, Salt Lake City, UT, USA, June 18-22, 2018. pp. 6077--6086. {IEEE} Computer Society (2018). \doi{10.1109/CVPR.2018.00636}

\bibitem{arjovsky2019invariant}
Arjovsky, M., Bottou, L., Gulrajani, I., Lopez-Paz, D.: Invariant risk minimization. arXiv preprint arXiv:1907.02893  (2019)

\bibitem{bolukbasi2016man}
Bolukbasi, T., Chang, K., Zou, J.Y., Saligrama, V., Kalai, A.T.: Man is to computer programmer as woman is to homemaker? debiasing word embeddings. In: Lee, D.D., Sugiyama, M., von Luxburg, U., Guyon, I., Garnett, R. (eds.) Advances in Neural Information Processing Systems 29: Annual Conference on Neural Information Processing Systems 2016, December 5-10, 2016, Barcelona, Spain. pp. 4349--4357 (2016)

\bibitem{cadene2019rubi}
Cad{\`{e}}ne, R., Dancette, C., Ben{-}younes, H., Cord, M., Parikh, D.: {RUBi}: {R}educing {U}nimodal {B}iases for {V}isual {Q}uestion {A}nswering. In: Wallach, H.M., Larochelle, H., Beygelzimer, A., d'Alch{\'{e}}{-}Buc, F., Fox, E.B., Garnett, R. (eds.) Advances in Neural Information Processing Systems 32: Annual Conference on Neural Information Processing Systems 2019, NeurIPS 2019, December 8-14, 2019, Vancouver, BC, Canada. pp. 839--850 (2019)

\bibitem{chawla2002smote}
Chawla, N.V., Bowyer, K.W., Hall, L.O., Kegelmeyer, W.P.: Smote: synthetic minority over-sampling technique. Journal of artificial intelligence research  \textbf{16},  321--357 (2002)

\bibitem{chen2020counterfactual}
Chen, L., Yan, X., Xiao, J., Zhang, H., Pu, S., Zhuang, Y.: Counterfactual samples synthesizing for robust visual question answering. In: Proceedings of the IEEE/CVF Conference on Computer Vision and Pattern Recognition. pp. 10800--10809 (2020)

\bibitem{chen2020learning}
Chen, X., Dai, H., Li, Y., Gao, X., Song, L.: Learning to stop while learning to predict. In: International Conference on Machine Learning. pp. 1520--1530. PMLR (2020)

\bibitem{choe2020empirical}
Choe, Y.J., Ham, J., Park, K.: An empirical study of invariant risk minimization. ICML 2020 Workshop on Uncertainty and Robustness in Deep Learning  (2020)

\bibitem{clark2019don}
Clark, C., Yatskar, M., Zettlemoyer, L.: Don{'}t take the easy way out: Ensemble based methods for avoiding known dataset biases. In: Proceedings of the 2019 Conference on Empirical Methods in Natural Language Processing and the 9th International Joint Conference on Natural Language Processing (EMNLP-IJCNLP). pp. 4069--4082. Association for Computational Linguistics, Hong Kong, China (2019). \doi{10.18653/v1/D19-1418}

\bibitem{clark2020learning}
Clark, C., Yatskar, M., Zettlemoyer, L.: Learning to model and ignore dataset bias with mixed capacity ensembles. In: Findings of the Association for Computational Linguistics: EMNLP 2020. pp. 3031--3045. Association for Computational Linguistics, Online (2020). \doi{10.18653/v1/2020.findings-emnlp.272}

\bibitem{creager2021environment}
Creager, E., Jacobsen, J.H., Zemel, R.: Environment inference for invariant learning. In: International Conference on Machine Learning. pp. 2189--2200. PMLR (2021)

\bibitem{cui2019class}
Cui, Y., Jia, M., Lin, T., Song, Y., Belongie, S.J.: Class-balanced loss based on effective number of samples. In: {IEEE} Conference on Computer Vision and Pattern Recognition, {CVPR} 2019, Long Beach, CA, USA, June 16-20, 2019. pp. 9268--9277. Computer Vision Foundation / {IEEE} (2019). \doi{10.1109/CVPR.2019.00949}

\bibitem{deng2009imagenet}
Deng, J., Dong, W., Socher, R., Li, L.J., Li, K., Fei-Fei, L.: Imagenet: A large-scale hierarchical image database. In: 2009 IEEE conference on computer vision and pattern recognition. pp. 248--255. Ieee (2009)

\bibitem{dosovitskiy2020image}
Dosovitskiy, A., Beyer, L., Kolesnikov, A., Weissenborn, D., Zhai, X., Unterthiner, T., Dehghani, M., Minderer, M., Heigold, G., Gelly, S., et~al.: An image is worth 16x16 words: Transformers for image recognition at scale. arXiv preprint arXiv:2010.11929  (2020)

\bibitem{duchi2019distributionally}
Duchi, J.C., Hashimoto, T., Namkoong, H.: Distributionally robust losses against mixture covariate shifts. Under review  (2019)

\bibitem{duggal2020elf}
Duggal, R., Freitas, S., Dhamnani, S., Horng, D., Sun, J., et~al.: Elf: An early-exiting framework for long-tailed classification. arXiv preprint arXiv:2006.11979  (2020)

\bibitem{grand2019adversarial}
Grand, G., Belinkov, Y.: Adversarial regularization for visual question answering: Strengths, shortcomings, and side effects. In: Proceedings of the Second Workshop on Shortcomings in Vision and Language. pp. 1--13. Association for Computational Linguistics, Minneapolis, Minnesota (2019). \doi{10.18653/v1/W19-1801}

\bibitem{guo2021attention}
Guo, M.H., Xu, T.X., Liu, J.J., Liu, Z.N., Jiang, P.T., Mu, T.J., Zhang, S.H., Martin, R.R., Cheng, M.M., Hu, S.M.: Attention mechanisms in computer vision: A survey. arXiv preprint arXiv:2111.07624  (2021)

\bibitem{he2008adasyn}
He, H., Bai, Y., Garcia, E.A., Li, S.: Adasyn: Adaptive synthetic sampling approach for imbalanced learning. In: 2008 IEEE international joint conference on neural networks (IEEE world congress on computational intelligence). pp. 1322--1328. IEEE (2008)

\bibitem{he2009learning}
He, H., Garcia, E.A.: Learning from imbalanced data. IEEE Transactions on knowledge and data engineering  \textbf{21}(9),  1263--1284 (2009)

\bibitem{he2017mask}
He, K., Gkioxari, G., Doll{\'a}r, P., Girshick, R.: Mask r-cnn. In: Proceedings of the IEEE international conference on computer vision. pp. 2961--2969 (2017)

\bibitem{he2016deep}
He, K., Zhang, X., Ren, S., Sun, J.: Deep residual learning for image recognition. In: Proceedings of the IEEE conference on computer vision and pattern recognition. pp. 770--778 (2016)

\bibitem{hettinger2017forward}
Hettinger, C., Christensen, T., Ehlert, B., Humpherys, J., Jarvis, T., Wade, S.: Forward thinking: Building and training neural networks one layer at a time. arXiv preprint arXiv:1706.02480  (2017)

\bibitem{hooker2020characterising}
Hooker, S., Moorosi, N., Clark, G., Bengio, S., Denton, E.: Characterising bias in compressed models. arXiv preprint arXiv:2010.03058  (2020)

\bibitem{howard2019searching}
Howard, A., Sandler, M., Chu, G., Chen, L.C., Chen, B., Tan, M., Wang, W., Zhu, Y., Pang, R., Vasudevan, V., et~al.: Searching for mobilenetv3. In: Proceedings of the IEEE/CVF International Conference on Computer Vision. pp. 1314--1324 (2019)

\bibitem{howard2017mobilenets}
Howard, A.G., Zhu, M., Chen, B., Kalenichenko, D., Wang, W., Weyand, T., Andreetto, M., Adam, H.: Mobilenets: Efficient convolutional neural networks for mobile vision applications. arXiv preprint arXiv:1704.04861  (2017)

\bibitem{hu2018gather}
Hu, J., Shen, L., Albanie, S., Sun, G., Vedaldi, A.: Gather-excite: Exploiting feature context in convolutional neural networks. Advances in neural information processing systems  \textbf{31} (2018)

\bibitem{hu2020triple}
Hu, T.K., Chen, T., Wang, H., Wang, Z.: Triple wins: Boosting accuracy, robustness and efficiency together by enabling input-adaptive inference. arXiv preprint arXiv:2002.10025  (2020)

\bibitem{huang2017densely}
Huang, G., Liu, Z., Van Der~Maaten, L., Weinberger, K.Q.: Densely connected convolutional networks. In: Proceedings of the IEEE conference on computer vision and pattern recognition. pp. 4700--4708 (2017)

\bibitem{jaderberg2015spatial}
Jaderberg, M., Simonyan, K., Zisserman, A., et~al.: Spatial transformer networks. Advances in neural information processing systems  \textbf{28} (2015)

\bibitem{kim2019learning}
Kim, B., Kim, H., Kim, K., Kim, S., Kim, J.: Learning not to learn: Training deep neural networks with biased data. In: {IEEE} Conference on Computer Vision and Pattern Recognition, {CVPR} 2019, Long Beach, CA, USA, June 16-20, 2019. pp. 9012--9020. Computer Vision Foundation / {IEEE} (2019). \doi{10.1109/CVPR.2019.00922}

\bibitem{kim2021biaswap}
Kim, E., Lee, J., Choo, J.: Biaswap: Removing dataset bias with bias-tailored swapping augmentation. In: Proceedings of the IEEE/CVF International Conference on Computer Vision. pp. 14992--15001 (2021)

\bibitem{krueger2021out}
Krueger, D., Caballero, E., Jacobsen, J.H., Zhang, A., Binas, J., Zhang, D., Le~Priol, R., Courville, A.: Out-of-distribution generalization via risk extrapolation (rex). In: International Conference on Machine Learning. pp. 5815--5826. PMLR (2021)

\bibitem{lee2016generalizing}
Lee, C.Y., Gallagher, P.W., Tu, Z.: Generalizing pooling functions in convolutional neural networks: Mixed, gated, and tree. In: Artificial intelligence and statistics. pp. 464--472. PMLR (2016)

\bibitem{lee2015deeply}
Lee, C.Y., Xie, S., Gallagher, P., Zhang, Z., Tu, Z.: Deeply-supervised nets. In: Artificial intelligence and statistics. pp. 562--570. PMLR (2015)

\bibitem{lee2022diversify}
Lee, Y., Yao, H., Finn, C.: Diversify and disambiguate: Learning from underspecified data. arXiv preprint arXiv:2202.03418  (2022)

\bibitem{li2019repair}
Li, Y., Vasconcelos, N.: {REPAIR:} removing representation bias by dataset resampling. In: {IEEE} Conference on Computer Vision and Pattern Recognition, {CVPR} 2019, Long Beach, CA, USA, June 16-20, 2019. pp. 9572--9581. Computer Vision Foundation / {IEEE} (2019). \doi{10.1109/CVPR.2019.00980}

\bibitem{lin2014microsoft}
Lin, T.Y., Maire, M., Belongie, S., Hays, J., Perona, P., Ramanan, D., Doll{\'a}r, P., Zitnick, C.L.: Microsoft coco: Common objects in context. In: European conference on computer vision. pp. 740--755. Springer (2014)

\bibitem{liu2021just}
Liu, E.Z., Haghgoo, B., Chen, A.S., Raghunathan, A., Koh, P.W., Sagawa, S., Liang, P., Finn, C.: Just train twice: Improving group robustness without training group information. In: International Conference on Machine Learning. pp. 6781--6792. PMLR (2021)

\bibitem{liu2021swin}
Liu, Z., Lin, Y., Cao, Y., Hu, H., Wei, Y., Zhang, Z., Lin, S., Guo, B.: Swin transformer: Hierarchical vision transformer using shifted windows. In: Proceedings of the IEEE/CVF International Conference on Computer Vision. pp. 10012--10022 (2021)

\bibitem{long2015fully}
Long, J., Shelhamer, E., Darrell, T.: Fully convolutional networks for semantic segmentation. In: Proceedings of the IEEE conference on computer vision and pattern recognition. pp. 3431--3440 (2015)

\bibitem{mehrabi2021survey}
Mehrabi, N., Morstatter, F., Saxena, N., Lerman, K., Galstyan, A.: A survey on bias and fairness in machine learning. ACM Computing Surveys (CSUR)  \textbf{54}(6),  1--35 (2021)

\bibitem{mostafa2018deep}
Mostafa, H., Ramesh, V., Cauwenberghs, G.: Deep supervised learning using local errors. Frontiers in neuroscience p.~608 (2018)

\bibitem{nam2020learning}
Nam, J., Cha, H., Ahn, S., Lee, J., Shin, J.: Learning from failure: Training debiased classifier from biased classifier. In: Advances in Neural Information Processing Systems (2020)

\bibitem{namkoong2016stochastic}
Namkoong, H., Duchi, J.C.: Stochastic gradient methods for distributionally robust optimization with f-divergences. In: Lee, D.D., Sugiyama, M., von Luxburg, U., Guyon, I., Garnett, R. (eds.) Advances in Neural Information Processing Systems 29: Annual Conference on Neural Information Processing Systems 2016, December 5-10, 2016, Barcelona, Spain. pp. 2208--2216 (2016)

\bibitem{nokland2016direct}
N{\o}kland, A.: Direct feedback alignment provides learning in deep neural networks. Advances in neural information processing systems  \textbf{29} (2016)

\bibitem{nokland2019training}
N{\o}kland, A., Eidnes, L.H.: Training neural networks with local error signals. In: International conference on machine learning. pp. 4839--4850. PMLR (2019)

\bibitem{pezeshki2020gradient}
Pezeshki, M., Kaba, S.O., Bengio, Y., Courville, A., Precup, D., Lajoie, G.: Gradient starvation: A learning proclivity in neural networks. arXiv preprint arXiv:2011.09468  (2020)

\bibitem{rahimian2019distributionally}
Rahimian, H., Mehrotra, S.: Distributionally robust optimization: A review. arXiv preprint arXiv:1908.05659  (2019)

\bibitem{ramakrishnan2018overcoming}
Ramakrishnan, S., Agrawal, A., Lee, S.: Overcoming language priors in visual question answering with adversarial regularization. In: Bengio, S., Wallach, H.M., Larochelle, H., Grauman, K., Cesa{-}Bianchi, N., Garnett, R. (eds.) Advances in Neural Information Processing Systems 31: Annual Conference on Neural Information Processing Systems 2018, NeurIPS 2018, December 3-8, 2018, Montr{\'{e}}al, Canada. pp. 1548--1558 (2018)

\bibitem{ronneberger2015u}
Ronneberger, O., Fischer, P., Brox, T.: U-net: Convolutional networks for biomedical image segmentation. In: International Conference on Medical image computing and computer-assisted intervention. pp. 234--241. Springer (2015)

\bibitem{DBLP:journals/corr/abs-1911-08731}
Sagawa, S., Koh, P.W., Hashimoto, T.B., Liang, P.: Distributionally robust neural networks for group shifts: On the importance of regularization for worst-case generalization. CoRR  \textbf{abs/1911.08731} (2019), \url{http://arxiv.org/abs/1911.08731}

\bibitem{sagawa2019distributionally}
Sagawa, S., Koh, P.W., Hashimoto, T.B., Liang, P.: Distributionally robust neural networks for group shifts: On the importance of regularization for worst-case generalization. arXiv preprint arXiv:1911.08731  (2019)

\bibitem{sagawa2020investigation}
Sagawa, S., Raghunathan, A., Koh, P.W., Liang, P.: An investigation of why overparameterization exacerbates spurious correlations. In: Proceedings of the 37th International Conference on Machine Learning, {ICML} 2020, 13-18 July 2020, Virtual Event. Proceedings of Machine Learning Research, vol.~119, pp. 8346--8356. {PMLR} (2020)

\bibitem{sanh2020learning}
Sanh, V., Wolf, T., Belinkov, Y., Rush, A.M.: Learning from others' mistakes: Avoiding dataset biases without modeling them. arXiv preprint arXiv:2012.01300  (2020)

\bibitem{scardapane2020should}
Scardapane, S., Scarpiniti, M., Baccarelli, E., Uncini, A.: Why should we add early exits to neural networks? Cognitive Computation  \textbf{12}(5),  954--966 (2020)

\bibitem{selvaraju2017grad}
Selvaraju, R.R., Cogswell, M., Das, A., Vedantam, R., Parikh, D., Batra, D.: Grad-cam: Visual explanations from deep networks via gradient-based localization. In: Proceedings of the IEEE international conference on computer vision. pp. 618--626 (2017)

\bibitem{shah2020pitfalls}
Shah, H., Tamuly, K., Raghunathan, A., Jain, P., Netrapalli, P.: The pitfalls of simplicity bias in neural networks. Advances in Neural Information Processing Systems  \textbf{33} (2020)

\bibitem{shrestha2022investigation}
Shrestha, R., Kafle, K., Kanan, C.: An investigation of critical issues in bias mitigation techniques. In: Proceedings of the IEEE/CVF Winter Conference on Applications of Computer Vision. pp. 1943--1954 (2022)

\bibitem{singh2020don}
Singh, K.K., Mahajan, D., Grauman, K., Lee, Y.J., Feiszli, M., Ghadiyaram, D.: Don't judge an object by its context: Learning to overcome contextual bias. In: 2020 {IEEE/CVF} Conference on Computer Vision and Pattern Recognition, {CVPR} 2020, Seattle, WA, USA, June 13-19, 2020. pp. 11067--11075. {IEEE} (2020). \doi{10.1109/CVPR42600.2020.01108}

\bibitem{szegedy2015going}
Szegedy, C., Liu, W., Jia, Y., Sermanet, P., Reed, S., Anguelov, D., Erhan, D., Vanhoucke, V., Rabinovich, A.: Going deeper with convolutions. In: Proceedings of the IEEE conference on computer vision and pattern recognition. pp.~1--9 (2015)

\bibitem{tan2019efficientnet}
Tan, M., Le, Q.: Efficientnet: Rethinking model scaling for convolutional neural networks. In: International conference on machine learning. pp. 6105--6114. PMLR (2019)

\bibitem{teerapittayanon2016branchynet}
Teerapittayanon, S., McDanel, B., Kung, H.T.: Branchynet: Fast inference via early exiting from deep neural networks. In: 2016 23rd International Conference on Pattern Recognition (ICPR). pp. 2464--2469. IEEE (2016)

\bibitem{teney2020unshuffling}
Teney, D., Abbasnejad, E., Hengel, A.v.d.: Unshuffling data for improved generalization. arXiv preprint arXiv:2002.11894  (2020)

\bibitem{teney2021evading}
Teney, D., Abbasnejad, E., Lucey, S., van~den Hengel, A.: Evading the simplicity bias: Training a diverse set of models discovers solutions with superior ood generalization. arXiv preprint arXiv:2105.05612  (2021)

\bibitem{tolstikhin2021mlp}
Tolstikhin, I.O., Houlsby, N., Kolesnikov, A., Beyer, L., Zhai, X., Unterthiner, T., Yung, J., Steiner, A., Keysers, D., Uszkoreit, J., et~al.: Mlp-mixer: An all-mlp architecture for vision. Advances in Neural Information Processing Systems  \textbf{34} (2021)

\bibitem{torralba2011unbiased}
Torralba, A., Efros, A.A.: Unbiased look at dataset bias. In: The 24th {IEEE} Conference on Computer Vision and Pattern Recognition, {CVPR} 2011, Colorado Springs, CO, USA, 20-25 June 2011. pp. 1521--1528. {IEEE} Computer Society (2011). \doi{10.1109/CVPR.2011.5995347}

\bibitem{utama2020towards}
Utama, P.A., Moosavi, N.S., Gurevych, I.: Towards debiasing {NLU} models from unknown biases. In: Proceedings of the 2020 Conference on Empirical Methods in Natural Language Processing (EMNLP). pp. 7597--7610. Association for Computational Linguistics, Online (2020). \doi{10.18653/v1/2020.emnlp-main.613}, \url{https://www.aclweb.org/anthology/2020.emnlp-main.613}

\bibitem{anonymous2020towards}
Utama, P.A., Moosavi, N.S., Gurevych, I.: Towards debiasing {NLU} models from unknown biases. In: Proceedings of the 2020 Conference on Empirical Methods in Natural Language Processing (EMNLP). pp. 7597--7610. Association for Computational Linguistics, Online (2020). \doi{10.18653/v1/2020.emnlp-main.613}, \url{https://www.aclweb.org/anthology/2020.emnlp-main.613}

\bibitem{venkataramani2015scalable}
Venkataramani, S., Raghunathan, A., Liu, J., Shoaib, M.: Scalable-effort classifiers for energy-efficient machine learning. In: Proceedings of the 52nd Annual Design Automation Conference. pp.~1--6 (2015)

\bibitem{wolczyk2021zero}
Wo{\l}czyk, M., W{\'o}jcik, B., Ba{\l}azy, K., Podolak, I., Tabor, J., {\'S}mieja, M., Trzcinski, T.: Zero time waste: Recycling predictions in early exit neural networks. Advances in Neural Information Processing Systems  \textbf{34} (2021)

\bibitem{xu2015show}
Xu, K., Ba, J., Kiros, R., Cho, K., Courville, A., Salakhudinov, R., Zemel, R., Bengio, Y.: Show, attend and tell: Neural image caption generation with visual attention. In: International conference on machine learning. pp. 2048--2057. PMLR (2015)

\bibitem{yu2021metaformer}
Yu, W., Luo, M., Zhou, P., Si, C., Zhou, Y., Wang, X., Feng, J., Yan, S.: Metaformer is actually what you need for vision. arXiv preprint arXiv:2111.11418  (2021)

\bibitem{zhang2018mitigating}
Zhang, B.H., Lemoine, B., Mitchell, M.: Mitigating unwanted biases with adversarial learning. In: Proceedings of the 2018 AAAI/ACM Conference on AI, Ethics, and Society. pp. 335--340 (2018)

\bibitem{zhou2017places}
Zhou, B., Lapedriza, A., Khosla, A., Oliva, A., Torralba, A.: Places: A 10 million image database for scene recognition. IEEE transactions on pattern analysis and machine intelligence  \textbf{40}(6),  1452--1464 (2017)

\bibitem{zhou2020bert}
Zhou, W., Xu, C., Ge, T., McAuley, J., Xu, K., Wei, F.: Bert loses patience: Fast and robust inference with early exit. Advances in Neural Information Processing Systems  \textbf{33},  18330--18341 (2020)

\end{thebibliography}
\clearpage
\renewcommand{\thepage}{A\arabic{page}}  
\renewcommand{\thesection}{A}   
\renewcommand{\thetable}{A\arabic{table}}   
\renewcommand{\thefigure}{A\arabic{figure}}

\section{Appendix}

\subsection{Detailed Results on Each Dataset}

\textbf{\gls*{BiasedMNIST}.} 
In Table~\ref{tab:biased-mnist-all-methods}, we present the unbiased test accuracies and majority/minority group accuracies for each bias variable. Methods run on Occam ResNet-18 lower the majority/minority discrepancy (MMD) compared to the methods run on ResNet-18 for all of the variables, indicating that OccamNets lower the tendencies to latch onto all of the spurious factors. OccamResNet-18 is especially robust to letter, letter color and texture color biases as shown by the low MMD values of $0 - 2.8\%$ compared to the larger MMD values obtained by ResNet-18. 
% \usepackage{multirow}
% \usepackage{graphicx}
\begin{table*}[h!]
\centering
\caption{Accuracies on majority (maj)/minority (min) groups for each bias variable in \gls*{BiasedMNIST} ($p_{bias}=0.95$).}
\label{tab:biased-mnist-all-methods}
%\addtolength\tabcolsep{2pt}
\resizebox{\textwidth}{!}{%
\begin{tabular}{lccccccccccccc} \hline
\multicolumn{1}{l}{\multirow{2}{*}{\begin{tabular}[c]{@{}l@{}}Architecture+Method\end{tabular}}} &
  \multicolumn{1}{l}{} &
  \multicolumn{2}{c}{\begin{tabular}[c]{@{}c@{}}Digit \\ Scale\end{tabular}} &
  \multicolumn{2}{c}{\begin{tabular}[c]{@{}c@{}}Digit\\ Color\end{tabular}} &
  \multicolumn{2}{c}{\begin{tabular}[c]{@{}c@{}}Texture\end{tabular}} &
  \multicolumn{2}{c}{\begin{tabular}[c]{@{}c@{}}Texture \\Color\end{tabular}} &
  \multicolumn{2}{c}{\begin{tabular}[c]{@{}c@{}}Letter\end{tabular}} &
  \multicolumn{2}{c}{\begin{tabular}[c]{@{}c@{}}Letter \\ Color\end{tabular}} \\
\multicolumn{1}{l}{} &
  \multicolumn{1}{l}{\begin{tabular}[c]{@{}l@{}}Test \\ Acc.\end{tabular}} &
  \multicolumn{2}{c}{maj/min} &
  \multicolumn{2}{c}{maj/min} &
  \multicolumn{2}{c}{maj/min} &
  \multicolumn{2}{c}{maj/min} &
  \multicolumn{2}{c}{maj/min} &
  \multicolumn{2}{c}{maj/min}
  \\ \hline
  & 
  \multicolumn{7}{c}{\textit{Results on ResNet-18}}  \\
 \multicolumn{1}{l}{ResNet+ERM} &
  \multicolumn{1}{l}{36.8} &
  \multicolumn{2}{c}{87.2/31.3} &
  \multicolumn{2}{c}{78.5/32.1} &
  \multicolumn{2}{c}{76.1/32.4} &
  \multicolumn{2}{c}{41.9/36.3} &
  \multicolumn{2}{c}{46.7/35.7} &
  \multicolumn{2}{c}{45.7/35.9} \\
  \multicolumn{1}{l}{ResNet+SD~\cite{pezeshki2020gradient}} &
  \multicolumn{1}{l}{37.1} &
  \multicolumn{2}{c}{83.4/32.0} &
  \multicolumn{2}{c}{76.9/32.7} &
  \multicolumn{2}{c}{76.7/32.7} &
  \multicolumn{2}{c}{42.3/36.6} &
  \multicolumn{2}{c}{48.3/35.8} &
  \multicolumn{2}{c}{48.9/35.9} \\
  \multicolumn{1}{l}{ResNet+UpWt} &
  \multicolumn{1}{l}{37.7} &
  \multicolumn{2}{c}{88.0/32.1} &
  \multicolumn{2}{c}{80.4/32.9} &
  \multicolumn{2}{c}{75.6/33.4} &
  \multicolumn{2}{c}{41.9/37.2} &
  \multicolumn{2}{c}{46.7/36.6} &
  \multicolumn{2}{c}{46.9/36.7} \\
  \multicolumn{1}{l}{ResNet+gDRO~\cite{sagawa2019distributionally}} &
  \multicolumn{1}{l}{19.2} &
  \multicolumn{2}{c}{55.0/15.2} &
  \multicolumn{2}{c}{50.2/15.7} &
  \multicolumn{2}{c}{63.4/14.2} &
  \multicolumn{2}{c}{24.8/18.6} &
  \multicolumn{2}{c}{26.7/18.3} &
  \multicolumn{2}{c}{29.5/18.1} \\
  \multicolumn{1}{l}{ResNet+PGI~\cite{ahmed2020systematic}} &
  \multicolumn{1}{l}{48.6} &
  \multicolumn{2}{c}{91.9/43.8} &
  \multicolumn{2}{c}{84.8/44.6} &
  \multicolumn{2}{c}{79.5/45.1} &
  \multicolumn{2}{c}{51.3/48.3} &
  \multicolumn{2}{c}{67.2/46.5} &
  \multicolumn{2}{c}{55.8/47.9} \\\hline
  & 
  \multicolumn{7}{c}{\textit{Results on OccamResNet-18}}  \\
  \multicolumn{1}{l}{OccamResNet} &
  \multicolumn{1}{l}{65.0} &
  \multicolumn{2}{c}{94.6/61.7} &
  \multicolumn{2}{c}{96.3/61.5} &
  \multicolumn{2}{c}{81.6/63.1} &
  \multicolumn{2}{c}{66.8/64.8} &
  \multicolumn{2}{c}{64.7/65.1} &
  \multicolumn{2}{c}{64.7/65.1} \\
    \multicolumn{1}{l}{OccamResNet+SD~\cite{pezeshki2020gradient}} &
  \multicolumn{1}{l}{55.2} &
  \multicolumn{2}{c}{92.3/51.1} &
  \multicolumn{2}{c}{92.9/50.9} &
  \multicolumn{2}{c}{78.9/52.5} &
  \multicolumn{2}{c}{57.4/54.9} &
  \multicolumn{2}{c}{55.9/55.1} &
  \multicolumn{2}{c}{55.3/55.2} \\
    \multicolumn{1}{l}{OccamResNet+UpWt} &
  \multicolumn{1}{l}{65.7} &
  \multicolumn{2}{c}{95.1/62.5} &
  \multicolumn{2}{c}{96.3/62.3} &
  \multicolumn{2}{c}{82.4/63.9} &
  \multicolumn{2}{c}{68.3/65.5} &
  \multicolumn{2}{c}{65.3/65.8} &
  \multicolumn{2}{c}{65.3/65.8} \\
  \multicolumn{1}{l}{OccamResNet+gDRO~\cite{sagawa2019distributionally}} &
  \multicolumn{1}{l}{29.8} &
  \multicolumn{2}{c}{72.8/25.0} &
  \multicolumn{2}{c}{69.5/25.3} &
  \multicolumn{2}{c}{45.8/28.0} &
  \multicolumn{2}{c}{39.4/28.8} &
  \multicolumn{2}{c}{29.7/29.8} &
  \multicolumn{2}{c}{36.1/29.1} \\
  \multicolumn{1}{l}{OccamResNet+PGI~\cite{ahmed2020systematic}} &
  \multicolumn{1}{l}{69.6} &
  \multicolumn{2}{c}{95.4/66.7} &
  \multicolumn{2}{c}{97.0/66.5} &
  \multicolumn{2}{c}{88.6/67.4} &
  \multicolumn{2}{c}{71.4/69.4} &
  \multicolumn{2}{c}{69.6/69.6} &
  \multicolumn{2}{c}{70.5/69.5} \\\hline

\end{tabular}}
\end{table*}

\textbf{\gls*{COCO}.} 
In Table~\ref{tab:coco-on-places-all-methods}, we present the accuracies on each of the test splits of \gls*{COCO}, alongside the average precision for the anomaly detection task. As discussed in~\ref{sec:results}, methods run on OccamResNet-18 show improvements over the methods run on ResNet-18 on the shifted test splits and the anomaly detection task. Furthermore, while PGI run on ResNet-18 shows a large drop of 7.4\% on the in-distribution test split, methods (barring gDRO) run on OccamResNet-18 show smaller drops of $0.1 - 2.1\%$, indicating robustness to distributions consisting of the same or different biases as compared to the train distribution.
% \usepackage{graphicx}
\begin{table}[h!]
\centering
\caption{Accuracy on the three splits of \gls*{COCO}, alongside average precision for the anomaly detection task.}
\label{tab:coco-on-places-all-methods}
% \addtolength\tabcolsep{5pt}
\resizebox{\textwidth}{!}{%
\begin{tabular}{lcccc}
 \hline
Methods &
  \begin{tabular}[c]{@{}c@{}}In-\\ Distribution\end{tabular} &
  \begin{tabular}[c]{@{}c@{}}Unseen\\ Backgrounds\end{tabular} &
  \begin{tabular}[c]{@{}c@{}}Seen, but\\ Non-Spurious \\Backgrounds \end{tabular} &
  \begin{tabular}[c]{@{}c@{}}Anomaly \\ Detection\end{tabular} \\ \hline
\multicolumn{5}{c}{\textit{Results on ResNet-18}}                                                             \\
ResNet+ERM  & \textbf{84.9} \tiny{$\pm$ 0.5} & 53.2 \tiny{$\pm$0.7} & 35.6 \tiny{$\pm$1.0} & 20.1 \tiny{$\pm$1.5} \\
ResNet+SD~\cite{pezeshki2020gradient} & \textbf{\underline{85.3}} \tiny{$\pm$0.3} & 52.8 \tiny{$\pm$0.9} & 35.4 \tiny{$\pm$0.5} & 19.9 \tiny{$\pm$1.4}    \\
ResNet+UpWt & \textbf{84.9} \tiny{$\pm$0.6} & 52.3 \tiny{$\pm$ 0.7}  & 35.2 \tiny{$\pm$ 0.4} & 20.4 \tiny{$\pm$1.9}   \\
ResNet+\gls*{gDRO}~\cite{sagawa2019distributionally} & 81.6 \tiny{$\pm$0.7} & 49.3 \tiny{$\pm$1.3} & 35.3 \tiny{$\pm$0.1}  & 19.6 \tiny{$\pm$1.7}   \\
ResNet+\gls*{PGI}~\cite{ahmed2020systematic} & 77.5 \tiny{$\pm$0.6} & 52.8 \tiny{$\pm$0.7} & 42.7 \tiny{$\pm$0.6} &  20.6 \tiny{$\pm$2.1} \\ 
\hline
\multicolumn{5}{c}{\textit{Results on OccamResNet-18}}                             \\
OccamResNet & 84.0 \tiny{$\pm$1.0}  & \textbf{55.8} \tiny{$\pm$1.2} & \textbf{43.4} \tiny{$\pm$1.0} &  \textbf{\underline{22.3}} \tiny{$\pm$2.8}  \\
OccamResNet+\gls*{SD}~\cite{pezeshki2020gradient} & \underline{84.8} \tiny{$\pm$0.4} & \underline{55.3} \tiny{$\pm$0.5} & 39.4 \tiny{$\pm$0.6} &  20.3 \tiny{$\pm$1.0} \\	
OccamResNet+\gls*{UpWt} & 82.9 \tiny{$\pm$0.5} & \textbf{\underline{56.6}} \tiny{$\pm$1.0} & \underline{42.9} \tiny{$\pm$0.8} &  \underline{21.0} \tiny{$\pm$0.9}  \\
OccamResNet+\gls*{gDRO}~\cite{sagawa2019distributionally} & 78.6 \tiny{$\pm$0.7} & 50.7 \tiny{$\pm$2.0} & 40.7 \tiny{$\pm$ 1.5}	&  19.3 \tiny{$\pm$2.3} \\
OccamResNet+\gls*{PGI}~\cite{ahmed2020systematic} & 82.8 \tiny{$\pm$0.6} & 55.3 \tiny{$\pm$1.3} & \textbf{\underline{43.6}} \tiny{$\pm$ 0.6}	&  \textbf{21.6} \tiny{$\pm$1.6} \\\hline
\end{tabular}%
}
\end{table}

\textbf{\gls*{BAR}.}
First of all, BAR consists of only 1941 samples, so we pre-trained ResNet-18 and OccamResNet-18 on 100 classes of ImageNet (obtaining 92.6\% and 92.1\% top-5 accuracies respectively) before training on BAR. Without the pre-trained weights, BAR obtains 15-20\% lower test set accuracies for both ResNet and OccamResNet as compared to the results with pre-trained weights. Now, as shown in Table~\ref{tab:BAR-all-methods}, methods run on OccamResNet show gains in terms of the overall test set accuracies over the methods run on ResNet. The per-class standard deviations are larger (1.8-16.2\%) as compared to the standard deviations for the overall test set accuracies (0.7-2.4\%). That is, across the five different experiments run with different random seeds, the same methods run on the same architectures end up favoring different classes. We hypothesize that despite starting off from the same initial conditions i.e., the same pre-trained parameters, the randomness in the mini-batches drive the models to favor certain classes over the others. Tuning the optimizer e.g., switching to SGD, lowering the learning rates or increasing the weight decay can potentially help mitigate the unstable behavior.
% \usepackage{graphicx}
\begin{table}[h!]
\centering
\caption{Overall and per-class accuracies on \gls*{BAR}}
\label{tab:BAR-all-methods}
%\addtolength\tabcolsep{1pt}
\resizebox{1.0\textwidth}{!}{%
\begin{tabular}{llllllll}
 \hline
Methods &
  \begin{tabular}[c]{@{}c@{}}Overall\end{tabular} &
  \begin{tabular}[c]{@{}c@{}}Climbing\end{tabular} &
  \begin{tabular}[c]{@{}c@{}}Diving\end{tabular} &
  \begin{tabular}[c]{@{}c@{}}Fishing\end{tabular} &
  \begin{tabular}[c]{@{}c@{}}Pole\\Vaulting\end{tabular} & \begin{tabular}[c]{@{}c@{}}Racing\end{tabular} & \begin{tabular}[c]{@{}c@{}}Throwing\end{tabular}
   \\ \hline
\multicolumn{8}{c}{\textit{Results on ResNet-18}} \\
ResNet+ERM & 51.3~\tiny{$\pm1.9$} & \textbf{\underline{69.5}}~\tiny{$\pm7.5$}  & 29.2~\tiny{$\pm1.8$} & 39.9~\tiny{$\pm16.2$} & 55.5~\tiny{$\pm6.4$} & \textbf{\underline{75.6}}~\tiny{$\pm5.6$} & \textbf{\underline{31.8}}~\tiny{$\pm4.3$} \\
ResNet+\gls*{SD}~\cite{pezeshki2020gradient} & 51.3~\tiny{$\pm2.3$} & \underline{62.1}~\tiny{$\pm7.5$} & 35.8~\tiny{$\pm2.0$} & 51.2~\tiny{$\pm6.4$} & 62.4~\tiny{$\pm9.2$} & 71.6~\tiny{$\pm10.0$} & 18.5~\tiny{$\pm6.7$} \\
ResNet+\gls*{UpWt} & 51.1~\tiny{$\pm1.9$} & 61.7~\tiny{$\pm13.2$} & \textbf{43.9}~\tiny{$\pm5.8$} & 42.3~\tiny{$\pm8.3$} & 52.3~\tiny{$\pm7.4$} & 67.9~\tiny{$\pm6.7$} & \textbf{28.2}~\tiny{$\pm12.8$} \\
ResNet+\gls*{gDRO}~\cite{sagawa2019distributionally} & 38.7~\tiny{$\pm2.2$} & 49.5~\tiny{$\pm8.5$} & 40.3~\tiny{$\pm8.4$} & 44.0~\tiny{$\pm10.4$} & 39.9~\tiny{$\pm7.1$} & 41.7~\tiny{$\pm4.0$} & 13.5~\tiny{$\pm5.9$} \\
ResNet+\gls*{PGI}~\cite{ahmed2020systematic} & \textbf{53.6}~\tiny{$\pm0.9$} & 61.2~\tiny{$\pm10.4$} & 38.4~\tiny{$\pm4.1$} & 42.9~\tiny{$\pm8.4$} & \textbf{\underline{73.3}}~\tiny{$\pm3.7$} & 68.9~\tiny{$\pm5.9$} & 23.5~\tiny{$\pm1.9$} \\ \hline
\multicolumn{8}{c}{\textit{Results on OccamResNet-18}} \\
OccamResNet & 52.6~\tiny{$\pm1.9$} & 59.3~\tiny{$\pm3.8$} & 42.3~\tiny{$\pm7.5$} & 44.6~\tiny{$\pm14.9$} & 60.5~\tiny{$\pm8.6$} & \underline{74.1}~\tiny{$\pm7.2$} & 22.1~\tiny{$\pm3.9$}  \\
OccamResNet+\gls*{SD}~\cite{pezeshki2020gradient} & 52.3~\tiny{$\pm2.4$} & 56.4~\tiny{$\pm6.8$} & 34.3~\tiny{$\pm5.8$} & \textbf{\underline{55.4}}~\tiny{$\pm7.4$} & \textbf{69.1}~\tiny{$\pm4.9$} & 72.9~\tiny{$\pm4.2$} & 21.8~\tiny{$\pm2.1$}  \\
OccamResNet+\gls*{UpWt} & 52.2~\tiny{$\pm1.4$} & 57.9~\tiny{$\pm1.8$} & 35.7~\tiny{$\pm7.5$} & \underline{51.8}~\tiny{$\pm11.2$} & 64.3~\tiny{$\pm8.8$} & 71.8~\tiny{$\pm3.8$} & \underline{27.4}~\tiny{$\pm3.5$} \\
OccamResNet+\gls*{gDRO}~\cite{sagawa2019distributionally} & \underline{52.9}~\tiny{$\pm0.8$} & 51.2~\tiny{$\pm9.6$} & \underline{42.8}~\tiny{$\pm8.2$} & 52.3~\tiny{$\pm5.1$} & 63.5~\tiny{$\pm7.3$} & \textbf{74.2}~\tiny{$\pm5.2$} & 25.3~\tiny{$\pm4.5$} \\
OccamResNet+\gls*{PGI}~\cite{ahmed2020systematic} & \textbf{\underline{55.9}}~\tiny{$\pm0.7$} & \textbf{64.2}~\tiny{$\pm5.1$} & \textbf{\underline{52.3}}~\tiny{$\pm6.4$} & 51.4~\tiny{$\pm8.3$} & \underline{64.4}~\tiny{$\pm4.1$} & 70.9~\tiny{$\pm8.1$} & 18.6~\tiny{$\pm6.8$} \\
\hline
\end{tabular}%
}
\end{table}

\subsection{Early Exit Statistics}
To examine the efficiency and robustness of each exit for all of the datasets, we present the exit \%, accuracy on the exited samples and overall exit-wise accuracies on all the samples for OccamResNet-18 in Table~\ref{tab:exit-stats}. For \gls*{BiasedMNIST}, the earliest exits $E_1$ and $E_2$ have high exit percentages of $59.8\%$ and $26.9\%$ respectively, alongside high accuracies on the exited samples: $68.1\%$ and $64.8\%$ respectively. These results show that OccamResNet has learned to identify and trigger earlier exits whenever appropriate. For \gls*{COCO}, we observe large accuracies of 50.8\% and 50.2\% on the 13.9\% and 49.6\% samples exited from $E_2$ and $E_3$ respectively. The large percentage of samples exiting from $E_3$ shows that OccamResNet is capable of using the full network depth whenever needed. The accuracy on the samples that exited from $E_1$ is however low: 31.3\%, even though the overall accuracy is 42.4\%, indicating need for improvement in terms of training the earlier exit gates. We believe that tuning the training thresholds more comprehensively can potentially close this gap. Finally, for \gls*{BAR}, more than half i.e., 55.9\% of the samples exit from $E_3$. The accuracies on the samples exited from $E_1$ and $E_2$: 55.0\% and 65.3\% are higher than the overall accuracies computed on all the samples i.e., 47.4\% and 52.3\% respectively. This again shows the ability to exit early whenever appropriate and the ability to utilize the full network depth only for the remaining samples.

% \usepackage{multirow}
% \usepackage{graphicx}
\begin{table*}[h!]
\centering
\caption{Percentage samples exited: (Exit \%), accuracy (Acc.) on exited samples and accuracy on all the samples for each exit ($E_j$).}
\label{tab:exit-stats}
\addtolength\tabcolsep{2pt}
% \resizebox{\textwidth}{!}{%
\begin{tabular}{lcccccccccccc} \hline
  \multicolumn{1}{l}{} &
  \multicolumn{4}{c}{\gls*{BiasedMNIST}} &
  \multicolumn{4}{c}{\gls*{COCO}} &
  \multicolumn{4}{c}{\gls*{BAR}} \\
\multicolumn{1}{l}{} &
  \multicolumn{1}{l}{$E_0$} &
  \multicolumn{1}{l}{$E_1$} &
  \multicolumn{1}{l}{$E_2$} &
  \multicolumn{1}{l}{$E_3$} &
  \multicolumn{1}{l}{$E_0$} &
  \multicolumn{1}{l}{$E_1$} &
  \multicolumn{1}{l}{$E_2$} &
  \multicolumn{1}{l}{$E_3$} &
  \multicolumn{1}{l}{$E_0$} &
  \multicolumn{1}{l}{$E_1$} &
  \multicolumn{1}{l}{$E_2$} &
  \multicolumn{1}{l}{$E_3$}
  \\ \hline
 Exit\%   & 0.0 & 59.8 & 26.9 & 13.3 & 0.0 & 36.5 & 13.9 & 49.6 & 0.0 & 23.7 & 20.3 & 55.9 \\
 Acc. (exited)   & N/A & 68.1 & 64.8 & 52.1 & N/A & 31.3 &  50.8 & 50.2 & N/A & 55.0 & 65.3 & 46.6 \\
 Acc. (all)   & 12.7 & 65.1 & 65.5 & 65.5 & 10.0 & 42.4 & 43.4 & 41.4 & 26.5 & 47.4 & 52.3 & 52.5 \\\hline
\end{tabular}
% }
\end{table*}

% \kk{What is the significance of shwoing accuracies here? I don't understand. If we only want to show exit percentages, we could consider sankey diagram which might look cooler..I do think it is very interesting how real images (COCO and BAR) would more likely require deeper processing --- we should highlight that...}\rs{Yeah, I added combo charts to show both exit\% and acc above (can make it look better once we decide on the right viz). I think acc is important as a sanity check that the samples exiting earlier indeed have reasonable accuracies.}

% \usepackage{multirow}
% \usepackage{graphicx}
\begin{table*}[h!]
\centering
\caption{Exit comparison on models trained with different levels of biases on \gls*{BiasedMNIST}. We present the percentage samples exited: (Exit \%) and accuracy on all the samples for each exit ($E_j$).}
\label{tab:bmnist-exit-stats}
\addtolength\tabcolsep{2pt}
% \resizebox{\textwidth}{!}{%
\begin{tabular}{lcccccccc} \hline
  \multicolumn{1}{l}{} &
  \multicolumn{4}{c}{$p_{bias}=0.5$} &
  \multicolumn{4}{c}{$p_{bias}=0.95$} \\
\multicolumn{1}{l}{} &
  \multicolumn{1}{l}{$E_0$} &
  \multicolumn{1}{l}{$E_1$} &
  \multicolumn{1}{l}{$E_2$} &
  \multicolumn{1}{l}{$E_3$} &
  \multicolumn{1}{l}{$E_0$} &
  \multicolumn{1}{l}{$E_1$} &
  \multicolumn{1}{l}{$E_2$} &
  \multicolumn{1}{l}{$E_3$}
  \\ \hline
 Exit\% & 0.0 & 67.3 & 23.8 & 8.9 & 0.0 & 53.6 & 34.2 & 12.2 \\
 Acc. (all) & N/A & 98.2 & 98.1 & 98.1 & N/A & 63.3 & 63.3 & 62.6 \\\hline
\end{tabular}
\end{table*}

% \usepackage{multirow}
% \usepackage{graphicx}
\begin{table*}[h!]
\centering
\caption{Exit comparison on different test splits of \gls*{COCO}. We present the percentage samples exited: (Exit \%), accuracy (Acc.) on exited samples and accuracy on all the samples for each exit ($E_j$).}
\label{tab:coco-exit-stats}
\addtolength\tabcolsep{2pt}
% \resizebox{\textwidth}{!}{%
\begin{tabular}{lcccccccccccc} \hline
  \multicolumn{1}{l}{} &
  \multicolumn{4}{c}{In-Distribution} &
  \multicolumn{4}{c}{Unseen Backgrounds} &
  \multicolumn{4}{c}{Unbiased Backgrounds} \\
\multicolumn{1}{l}{} &
  \multicolumn{1}{l}{$E_0$} &
  \multicolumn{1}{l}{$E_1$} &
  \multicolumn{1}{l}{$E_2$} &
  \multicolumn{1}{l}{$E_3$} &
  \multicolumn{1}{l}{$E_0$} &
  \multicolumn{1}{l}{$E_1$} &
  \multicolumn{1}{l}{$E_2$} &
  \multicolumn{1}{l}{$E_3$} &
  \multicolumn{1}{l}{$E_0$} &
  \multicolumn{1}{l}{$E_1$} &
  \multicolumn{1}{l}{$E_2$} &
  \multicolumn{1}{l}{$E_3$}
  \\ \hline
 Exit\%  & 0.0 & 54.6 & 12.4 & 33.0 & 0.0 & 13.1 & 9.7 & 77.2 & 0.0 & 39.4 & 14.5 & 46.1 \\
 Acc. (exited) & N/A & 94.7 & 86.5 & 66.9 & N/A & 58.9 & 67.8 & 52.3 & N/A & 27.5 & 55.4 & 49.2 \\
 Acc. (all) & 67.2 & 81.1 & 84.0 & 85.3 & 20.9 & 51.7 & 55.1 & 55.4 & 11.2 & 40.1 & 41.6 & 39.4 \\\hline
\end{tabular}
\end{table*}

\textbf{Exit Statistics on differently shifted distributions.}

In general, we find that earlier exits are triggered more often for in-distribution (easier) test samples as compared to shifted distribution (more difficult) test samples. As shown in Table~\ref{tab:bmnist-exit-stats}, for BiasedMNIST, when $p_{bias}$ is increased from $0.5$ (easy) to $0.95$ (hard), exit\% of the earliest exit: $E_1$ drops from 67.3\% to 53.6\%. Similarly, as shown in Table~\ref{tab:coco-exit-stats}, for COCO-on-Places, $E_1$'s exit\% on in-distribution (easy) split is 54.6\%, whereas it is 39.4\% on unbiased backgrounds (hard) split. However, for the test split with unseen backgrounds, $E_1$'s exit\%: 13.1\% is lower than the exit\% of 39.4\% obtained for the test split with unbiased backgrounds, despite the latter being more difficult. OccamNet failed to trigger earlier exits even though $E_1$ was accurate for 51.7\% of the samples $E_2$ was comparable with $E_3$ in terms of overall accuracy. We hypothesize that the earlier exits failed to trigger since they had not been trained to exit when the backgrounds are out-of-distribution. Thus, one area for improvement is to enable the ability to exit confidently in spite of the presence of previously unseen factors. Note that this analysis was performed on a single run, so may have small differences with the multi-run averages presented elsewhere.

\subsection{Using comparable \# of parameters}
In the main paper, we compared OccamResNet-18 with 8M parameters (feature width = 48) and ResNet-18 with 12M parameters (feature width = 64). To examine if the lower number number of parameters is helping e.g., due to implicit regularization, we test an OccamResNet-18 with 12M parameters by setting the feature width to 58. As shown in Table~\ref{tab:overall-12M}, OccamResNet-18 with 12M parameters shows small improvements over OccamResNet-18 with 8M parameters in all the datasets. A more thorough analysis of model sizes and their impacts on accuracy is an interesting study and we leave this to future work.
\begin{table}[t]
\centering
\footnotesize
\caption{We train on OccamResNet-18-width-58 with 12M parameters (feature width set to 58) to make the number of parameters comparable to ResNet-18 (12M parameters, feature width=48).}
\label{tab:overall-12M}
\addtolength\tabcolsep{5pt}
\begin{tabular}{lccc}
 \hline
Architecture+Method &
  \begin{tabular}[c]{@{}l@{}}\gls*{BiasedMNIST}\end{tabular} &
  \begin{tabular}[c]{@{}l@{}}\gls*{COCO}\end{tabular} &
  \begin{tabular}[c]{@{}l@{}}\gls*{BAR} \end{tabular} \\ \hline
\multicolumn{4}{c}{\textit{Results on Standard \gls*{ResNet}-18 (12M parameters, feature width=64)}}                                                             \\
ResNet+ERM  & 36.8 \tiny{$\pm$ 0.7} & 35.6 \tiny{$\pm$ 1.0}  & 51.3 \tiny{$\pm$1.9}    \\
ResNet+\gls*{SD}~\cite{pezeshki2020gradient} & 37.1 \tiny{$\pm$ 1.0} & 35.4 \tiny{$\pm$ 0.5}  & 51.3 \tiny{$\pm$2.3}    \\
ResNet+\gls*{UpWt} & 37.7 \tiny{$\pm$ 1.6} & 35.2 \tiny{$\pm$ 0.4}  & 51.1 \tiny{$\pm$1.9}   \\
ResNet+\gls*{gDRO}~\cite{sagawa2019distributionally} & 19.2 \tiny{$\pm$ 0.9} & 35.3 \tiny{$\pm 0.1$}  & 38.7 \tiny{$\pm$2.2}  \\
ResNet+\gls*{PGI}~\cite{ahmed2020systematic} & \textbf{48.6} \tiny{$\pm$ 0.7} & \textbf{42.7} \tiny{$\pm$ 0.6} & \textbf{53.6} \tiny{$\pm$0.9} \\ 
\hline
\multicolumn{4}{c}{\textit{Results on \gls*{OccamResNet}-18 (8M parameters, feature width=48)}}                                                             \\
\gls*{OccamResNet} & \textbf{65.0} \tiny{$\pm 1.0$} & \textbf{43.4} \tiny{$\pm$ 1.0} & \underline{52.6} \tiny{$\pm$1.9} \\
\hline
\multicolumn{4}{c}{\textit{Results on \gls*{OccamResNet}-18-width-58 (12M parameters, feature width=58)}}                                                       \\
\gls*{OccamResNet} & \textbf{\underline{65.9}} \tiny{$\pm 1.3$} & \textbf{\underline{43.8}} \tiny{$\pm$ 1.1} & \textbf{\underline{53.5}} \tiny{$\pm$2.2} \\
%\gls*{OccamResNet}+PGI & \textbf{\underline{69.6}} \tiny{$\pm 0.7$} & \textbf{\underline{43.6}} \tiny{$\pm$ 0.6} & \textbf{\underline{55.9}} \tiny{$\pm$0.7} \\
\hline
\end{tabular}%
\end{table}
% If there is some combination (ours+XXX) does really good, We could separate Ours into two lines, Ours: Occament and Ours:Final (Occamnet+PGI) perhaps...
%\clearpage

\subsection{Sample Complexity}

\begin{figure}
\footnotesize
 \centering
    \includegraphics[width=0.75\linewidth]{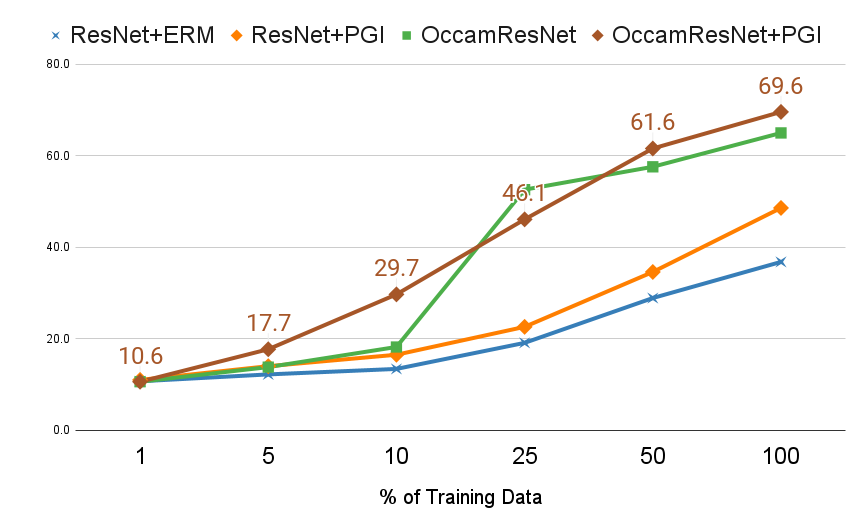}
      \caption{Unbiased accuracies obtained when trained with the indicated percent of training data.}
      \label{fig:bmnist-sample-complexity}
\end{figure}

It is desirable to have models that generalize despite being trained with a limited number of samples i.e., with reduced sample complexity. This is especially true for biased datasets, where reducing the train set size can amplify biases~\cite{anonymous2020towards}. To study the ability to generalize when only a subset of the training data is available, we train ResNet and OccamResNet on ${1\%, 5\%, 10\%, 25\%, 50\%, 100\%}$ of \gls*{BiasedMNIST}'s train set. As shown in Fig.~\ref{fig:bmnist-sample-complexity}, OccamResNet (without PGI) trained on only 25\% of the data outperforms ResNet+ERM and ResNet+PGI trained on 100\% of the data showing increased sample complexity. When trained on only 10\% of the training set, OccamResNet+PGI outperforms rest of the methods by large margins of $11.5-16.2\%$ showing that OccamResNet with group labels show the greatest efficacy in the low-shot data regime. When only 1\% of the training data is available, all the methods obtain chance-level accuracies (i.e., near 10\%) indicating lack of enough sufficient training samples for classification. For the rest, methods run on OccamResNet-18 outperform the methods run on ResNet-18, showing improved sample complexity.

% \begin{figure}[ht!]
% \footnotesize
%  \centering
%     \includegraphics[width=0.75\linewidth]{figures/bmnist-sample-complexity.png}
%       \caption{Unbiased accuracies obtained when trained with the indicated percent of training data.}
%       \label{fig:bmnist-sample-complexity}
% \end{figure}

\subsection{Robustness to Varying Levels of Bias in \gls*{BiasedMNIST}}
% \begin{wrapfigure}[14]{r}{0.5\textwidth}
%      \includegraphics[width=0.95\linewidth]{figures/bmnist-bias-levels.png}
%       \caption{Unbiased accuracies at varying bias levels ($p_{bias}$) in \gls*{BiasedMNIST}.}
%       \label{fig:bmnist-bias-levels}
% \end{wrapfigure}

\begin{figure}
 \centering
    \includegraphics[width=0.75\linewidth]{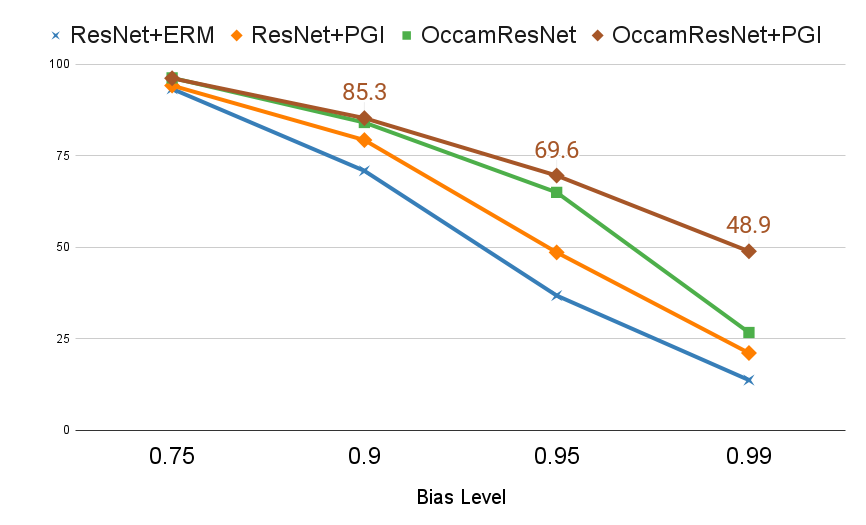}
      \caption{Unbiased accuracies at varying bias levels ($p_{bias}$) in \gls*{BiasedMNIST}.}
      \label{fig:bmnist-bias-levels}
\end{figure}

To gauge the robustness of models, it is important to examine their behaviors across varying levels of biases. For this, we present the unbiased accuracies obtained by training separate models on training sets with $p_{bias} \in \{0.75, 0.9, 0.95, 0.99\}$. As shown in fig.~\ref{fig:bmnist-bias-levels}, all of the methods obtain similar accuracies at $p_{bias} = 0.75$, where bias is not severe. OccamResNet+PGI outperform rest of the methods at $p_{bias} = \{0.9, 0.95, 0.99\}$. The gap between OccamResNet+PGI and other methods are especially drastic for $p_{bias} = 0.99$, indicating that when OccamResNet is trained to have similar prediction distributions across groups, it is capable of tackling highly biased training distributions too.

\subsection{Evaluation on Other Architectures}
\label{sec:other_archs}
Apart from ResNet, we also tested the proposed inductive biases on EfficientNet and MobileNet. The results are presented in Table~\ref{tab:other-archs}. For both \gls*{BiasedMNIST} and \gls*{COCO}, Occam variants outperform the standard architectures, showing the efficacy of the proposed modifications.
 
 \begin{table}[]
\centering
\footnotesize
\caption{Unbiased test set accuracies comparing the standard and Occam variants of ResNet-18, EfficientNet-B2 and MobileNetv3 architectures, run without additional debiasing procedures.}
\label{tab:other-archs}
\addtolength\tabcolsep{5pt}
\begin{tabular}{lccc}
 \hline
Architecture & \begin{tabular}[c]{@{}l@{}}Number of \\ Parameters\end{tabular} &
  \begin{tabular}[c]{@{}l@{}}\gls*{BiasedMNIST}\end{tabular} &
  \begin{tabular}[c]{@{}l@{}}\gls*{COCO}\end{tabular} \\ \hline
ResNet-18 & 12M &  36.8 & 35.6 \\
OccamResNet-18 & 12M & 65.9 & 43.8 \\
EfficientNet-B2 & 9M & 34.4 & 34.2 \\
OccamEfficientNet-B2 & 9M & 59.2 & 39.2 \\ 
MobileNet-v3 & 5.5M & 40.4 & 34.9 \\
OccamMobileNet-v3 & 5.5M & 49.9 & 40.1 \\
\hline
\end{tabular}%
\end{table}

\subsection{OccamNet Implementation Details}
\label{sec:arch_details}

In OccamNet, each exit module $E_j$ takes in feature maps produced by the corresponding block $B_j$ of the backbone network. $E_j$ consists of two $3\times3$ convolutional layers $(F_j)$ for the initial pre-processing of the feature maps. $F_j$ consists of convolutional layers with the number of channels set to: $max(\frac{d_j}{4}, d_{min})$, where $d_j$ is the number of channels in the feature maps produced by $B_j$, and $d_{min}$ is set to 32 for OccamResNet and OccamMobileNet and 16 for OccamEfficientNet. Feature maps from $F_j$ are fed into the CAM predictor $C_j$ and the exit gate $G_j$.  $C_j$ is a $1\times1$ convolutional layer with the number of output channels set to the number of classes, $n_Y$. $G_j$ consists of a 16-dimensional hidden ReLU layer followed by a sigmoid layer that predicts the exit probability.

\textbf{Exit Details.}
For convenience, we specify the exit locations with reference to PyTorch 1.7.1 implementations of the architectures. For ResNet, the residual layers that yield the same number of output channels are grouped together and we refer to each of those groups as a `block'. ResNet-18 consists of 4 blocks and we attach an exit to each of the blocks. For OccamResNet-18, with feature width of 58, the exit-wise input dimensions are: $E_0: 58$, $E_1: 116$, $E_2: 232$ and $E_3: 464$. Similarly, EfficientNet-B2 consists of 9 blocks and we attach the exits to the $3^{rd}, 5^{th}, 7^{th}~and~9^{th}$ blocks. We decrease the width multiplier of 1.1 in the standard architecture to 0.88 in OccamEfficientNet-B2 to create a model with comparable number of parameters of 9M for both. The input dimensions of the corresponding exits are: $E_0: 24$, $E_1: 72$, $E_2: 168$ and $E_3: 1120$. Finally, MobileNetv3-large consists of 17 blocks, and the exits are attached to the $2^{nd}, 7^{th}, 13^{th}$ and the $17^{th}$ blocks. We decrease the width multiplier from a value of $1$ in MobileNet-v3-large to 0.95 in OccamMobileNet-v3-large, so that both models have 5.5M parameters. The input dimensions of the corresponding exits are: $E_0: 16, E_1: 40, E_2: 104, E_3: 912$.  

\textbf{Modifications for \gls*{COCO}.}
For \gls*{COCO}, the images are small ($64 \times 64$), so for ResNet-18 and OccamResNet-18, we replace the first convolutional  layer (kernel size=$7$, padding=$3$, stride=$2$), with a smaller layer (kernel size=$3$, padding=$1$ and stride=$1$) and also remove the initial max pooling layer. For the standard and Occam variants of EfficientNet-B2 and MobileNet-v3, we scale up the image size to $224 \times 224$, which improved the accuracy.

\textbf{Computational Costs and Training Durations}
OccamResNet18 incurs additional multiply-accumulate (Mac) operations, requiring 2.11 GMacs compared to 1.82 GMacs required by ResNet18. While OccamResNet is slower than ERM on ResNet, it is faster than PGI run on ResNet. Average training durations per epoch for ERM on ResNet, PGI on ResNet and OccamResNet are: 10, 17, 14 secs for COCO-on-Places and 40, 67, 60 secs for BiasedMNIST on a Titan RTX.

\subsection{Hyperparameters and Other Settings}
\label{sec:hyperparams}
In Table~\ref{tab:hyperparameters}, we present the details about optimizers, training epochs and other hyperparameters for each method on each dataset. The hyperparameter search grids for OccamResNet-18 and all of the comparison methods are shown below. For each dataset, we tune ResNet-18 and OccamResNet-18 separately.

\textbf{Spectral Decoupling (SD).} The output decay term $\lambda$ is used to penalize the model predictions by using a regularizer: $\frac{\lambda}{2} ||\hat{y}||^2_2$. We search for the output decay term $\lambda \in \{1, 0.1, 0.01, 10^{-3}, 10^{-4}\}$.

\textbf{Group Upweighting (UpWt).} Group adjustment hyperparameter, i.e., the exponentiation factor $\gamma$ is used to balance the group-wise contributions $\frac{1}{n_g^\gamma}$, where $n_g$ is the number of samples in group $g$. We search for $\gamma \in \{0.5, 1, 2, 3\}$.

\textbf{Group DRO (gDRO).} Again, we search for the group adjustment hyperparameter, i.e., $\gamma \in \{0.5,1,2,3\}$. Group weight step size, which is used to control the group-wise loss weights is selected by searching from these values: $\{0.1, 0.01, 10^{-3}, 10^{-4}\}$. 

\textbf{Predictive Group Invariance (PGI).} The search range for the invariance penalty loss, i.e., the KLD loss between different groups from the same class is: \{1, 10, 50, 100, 500, 1000\}.

\textbf{OccamNets.} For OccamNets, we recommend tuning the accuracy threshold of the first exit: ($\tau_{acc,0}$) on a validation set, but fixing rest of the hyperparameters, based on the following observations:

\begin{itemize}
    \item \textit{Bias Amplification Factor ($\gamma_0$) and Weight offset ($\epsilon$):} We tuned $\gamma_0$ and $\epsilon$ on COCO-on-Places, but fixed the values for rest of the datasets. We observe that $\gamma_0 \ge 3$ ensures sufficient bias amplification, so recommend using $\gamma_0 = 3$. Furthermore, $\epsilon=0.1$ ensures non-zero losses in all the datasets, so we recommend using this default value.
    
    \item \textit{Accuracy Thresholds ($\tau_{acc}$):}  We use arithmetic progression for the mean-per-class accuracy thresholds $\tau_{acc}$, with the difference $\Delta \tau_{acc,j}$, set to 0.1, i.e., the threshold is increased by $0.1$ every subsequent exit. We search for the initial training threshold $\tau_{acc,0} \in \{0.1, 0.3, 0.5\}$. BMNIST and COCO were relatively insensitive to $\tau_{acc,0}$, with absolute differences of only: 1-2\% in accuracy and 1-4\% in exit\%. For BAR and ImageNet, we decreased $\tau_{acc,0}$ to $0.1$ since higher values increase exit\% of $E_1$, which decreases the overall accuracy. So, we recommend tuning $\tau_{acc,0}$.
    
    \item \textit{Normalization term:} The balancing/normalization in equation 4 can be generalized as $(\frac{1}{\sum_{1(g_j=k)}})^\beta$. We searched for $\beta$ in $\{0.5,1.0\}$. With $\beta=1$, only 8.8\% samples exited from $E_1$ for BMNIST, compared to 59.8\% with $\beta=0.5$. Accuracies for $\beta=$0.5/1.0 were similar: e.g., 65.0\%/64.2\% for BMNIST and 44.1\%/44.0\% for COCO (single runs), so we chose $\beta=0.5$ to favor earlier exits on all the datasets.
\end{itemize}

\begin{table*}[h!]
\centering
\caption{Hyperparameters and other settings used for each method on all of the datasets.}
\label{tab:hyperparameters}
\resizebox{1.0\textwidth}{!}{%
% \addtolength\tabcolsep{5pt}
\begin{tabular}{llccc}
\hline
{\begin{tabular}[c]{@{}l@{}}\textbf{Datasets/}\\\textbf{Methods} \end{tabular}} &
  {\begin{tabular}[c]{@{}l@{}}\textbf{Setting} \end{tabular}} &
  {\begin{tabular}[c]{@{}l@{}}\textbf{Biased}\\\textbf{MNIST} \end{tabular}} &
  {\begin{tabular}[c]{@{}l@{}}\textbf{COCO on}\\\textbf{Places} \end{tabular}} &
  \textbf{\gls*{BAR}} \\ \hline
  Common to all the methods & Optimizer & Adam & SGD & Adam \\ 
  & Learning Rate (LR) & $10^{-3}$ & 0.1 & 5e-4 \\
  & LR Decay Milestones & [50,70] & [100,120,140] & - \\
  & LR Decay Gamma & 0.1 & 0.1 & - \\
  & Weight Decay & $5 \times 10^{-4}$ &  $5 \times 10^{-4}$  & $5 \times 10^{-4}$ \\
  & Momentum & - & 0.9 & - \\
  & Batch Size & 128 & 64 & 128 \\
  & Epochs & 90 & 150 & 150 \\
\hline
\begin{tabular}[c]{@{}l@{}}Spectral Decoupling (SD) \\ on ResNet-18~\cite{pezeshki2020gradient} \end{tabular}  &
  Output Decay ($\lambda$) &
  $\lambda$ = 0.1
  & \begin{tabular}[c]{@{}c@{}}$\lambda_{min.} = 10^{-3}$ \\ $\lambda_{maj.} = 0.1$ \end{tabular} & $\lambda = 0.1$ \\
   \hline
  \begin{tabular}[c]{@{}l@{}}Spectral Decoupling (SD) \\ on OccamResNet-18~\cite{pezeshki2020gradient} \end{tabular}  &
  Output Decay ($\lambda$) &
  $\lambda$ = $10^{-3}$ & \begin{tabular}[c]{@{}c@{}}$\lambda_{min.} = 10^{-3}$ \\ $\lambda_{maj.} = 0.1$ \end{tabular} & $\lambda = 10^{-3}$
   \\  \hline
\gls*{UpWt} &
\begin{tabular}[c]{@{}l@{}}Exponentiation \\ Factor ($\gamma$) \end{tabular} &
  2 &
  1 &
  1 \\ \hline
\begin{tabular}[c]{@{}l@{}}Group DRO (gDRO) \\ on ResNet-18~\cite{sagawa2019distributionally} \end{tabular} &
  \begin{tabular}[c]{@{}l@{}}Step size \end{tabular} &
  $10^{-3}$ &
  $10^{-3}$ &
  $0.01$ \\
   &
  \begin{tabular}[c]{@{}l@{}}Exponentiation \\ Factor ($\gamma$) \end{tabular} &
  $0.5$ &
  $1$ &
  $0.01$ \\ \hline
\begin{tabular}[c]{@{}l@{}}Group DRO (gDRO) \\ on OccamResNet-18~\cite{sagawa2019distributionally} \end{tabular} &
  \begin{tabular}[c]{@{}l@{}}Step size \end{tabular} &
  $10^{-3}$ &
  $10^{-4}$ &
  $0.01$ \\
   &
  \begin{tabular}[c]{@{}l@{}}Exponentiation \\ Factor ($\gamma$) \end{tabular} &
  $0.5$ &
  $0.5$ &
  $0.5$ \\  
  \hline
 \begin{tabular}[c]{@{}l@{}}Predictive Group \\ Invariance (PGI) \\ on ResNet-18~\cite{ahmed2020systematic} \end{tabular} &
  \begin{tabular}[c]{@{}l@{}}Invariance Loss \\ Weight \end{tabular} & 100 & 50 & 10 \\ \hline
\begin{tabular}[c]{@{}l@{}}Predictive Group \\ Invariance (PGI) \\ on OccamResNet-18~\cite{ahmed2020systematic} \end{tabular} &
\begin{tabular}[c]{@{}l@{}}Invariance Loss \\ Weight \end{tabular} & 50 & 1 & 50 \\ \hline
OccamNets & \begin{tabular}[c]{@{}l@{}}Threshold \\ for $E_0$ ($\tau_{acc,0}$) \end{tabular} & 0.5 & 0.5 & 0.1 \\ 
& \begin{tabular}[c]{@{}l@{}}CAM Suppression \\ Loss Weight ($\lambda_{CS}$) \end{tabular} & 0.1 & 0.1 & 0.1 \\
\hline
\end{tabular}%
}
\end{table*}

\subsection{Issues Training with GroupDRO (gDRO)}
\label{sec:gdro}
We find that gDRO on \gls*{BiasedMNIST} and ResNet+gDRO on \gls*{BAR} obtain accuracies lower than ResNet+ERM. To alleviate this issue, we tried to tune the hyperparameters by lowering the learning rates to $\{10^{-4}, 10^{-5}\}$ and increasing the weight decays to $\{0.1, 0.01, 10^{-3}\}$ as suggested in~\cite{sagawa2019distributionally}, yet \gls*{gDRO} obtained low accuracies. We believe the challenge stems from the large number of dataset groups in \gls*{BiasedMNIST} and the small training set size of \gls*{BAR}. While optimizing \gls*{gDRO} on such conditions still remains a challenge, gDRO run on OccamResNet showed accuracy gains of 10.6\% on \gls*{BiasedMNIST} and 14.2\% on \gls{BAR} over gDRO run on ResNet, indicating that OccamNets also offer better training process.

\subsection{Augmentations}
\textbf{\gls*{BiasedMNIST}.} We do not perform any augmentation.

\textbf{\gls*{COCO}.} Following~\cite{ahmed2020systematic}, we apply random cropping by padding the original images by 8 pixels on all the sides (reflection padding) and taking $64\times64$ random crops. We also apply random horizontal flips.

\textbf{\gls*{BAR}.} We apply random resized crops using a scale range of 0.7 to 1.0 and selecting aspect ratios between 1.0 to $\frac{4}{3}$. We also apply random horizontal flips.

\subsection{Model Calibration}
\begin{figure}
\footnotesize
 \centering
    \includegraphics[width=0.24\linewidth]{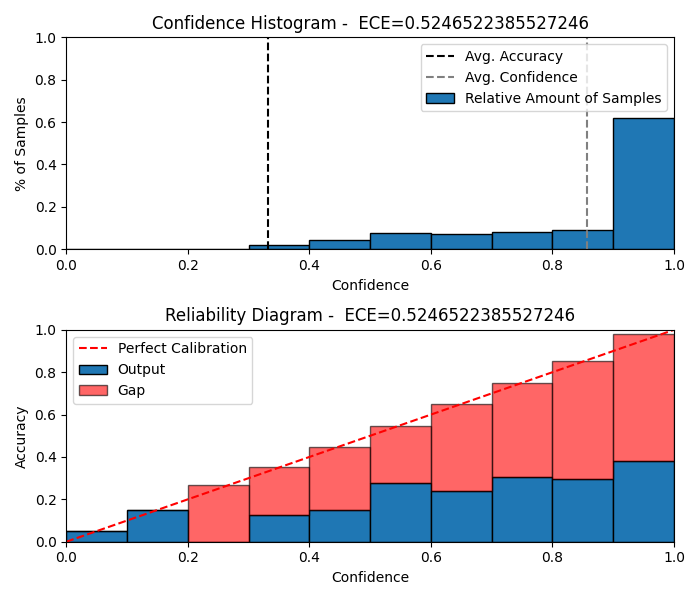}
    \includegraphics[width=0.24\linewidth]{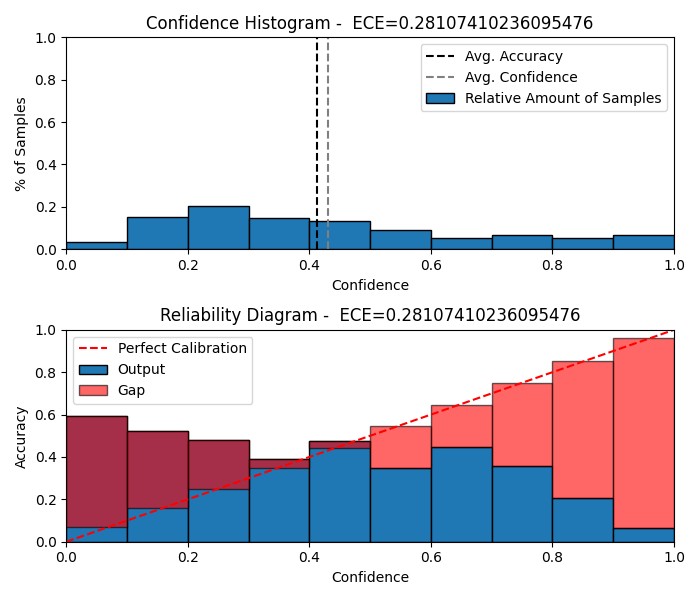}
    \includegraphics[width=0.24\linewidth]{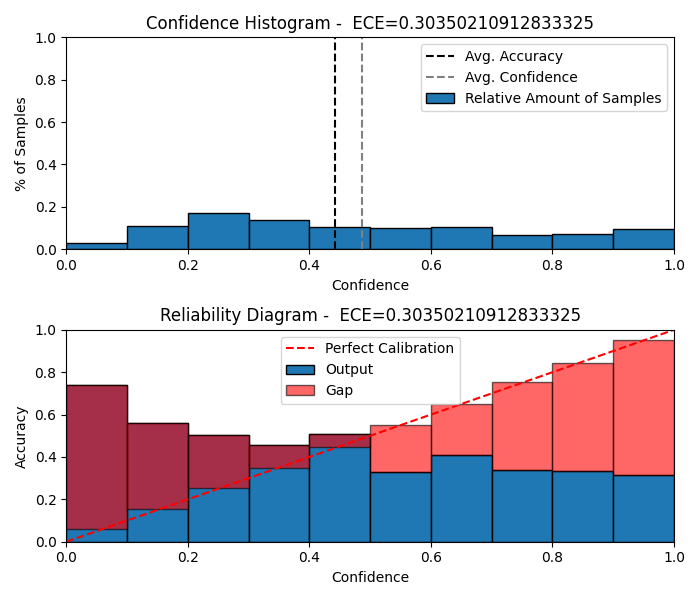}
    \includegraphics[width=0.24\linewidth]{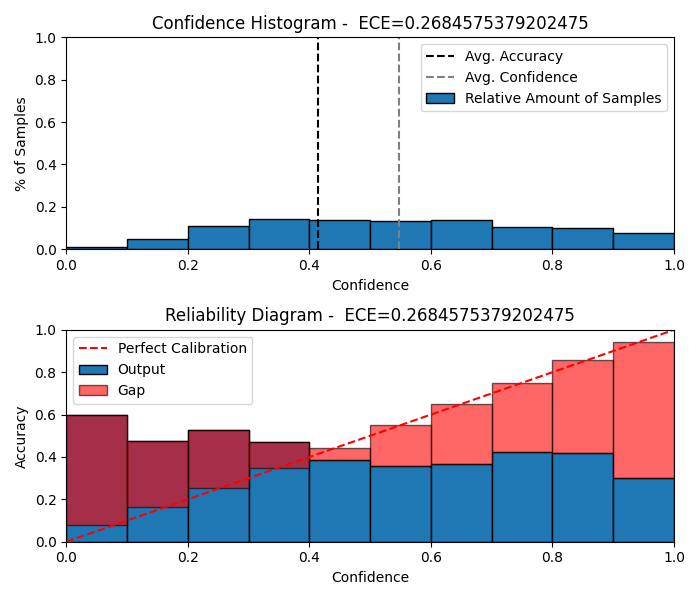}
      \caption{Reliability diagrams for the classifier trained with ERM (leftmost column) versus exit gate calibrations for $E1-E3$ (right hand columns) on \gls*{COCO} (unbiased backgrounds test split).}
      \label{fig:coco-calibration}
\end{figure}

\begin{figure}
\footnotesize
 \centering
    \includegraphics[width=0.24\linewidth]{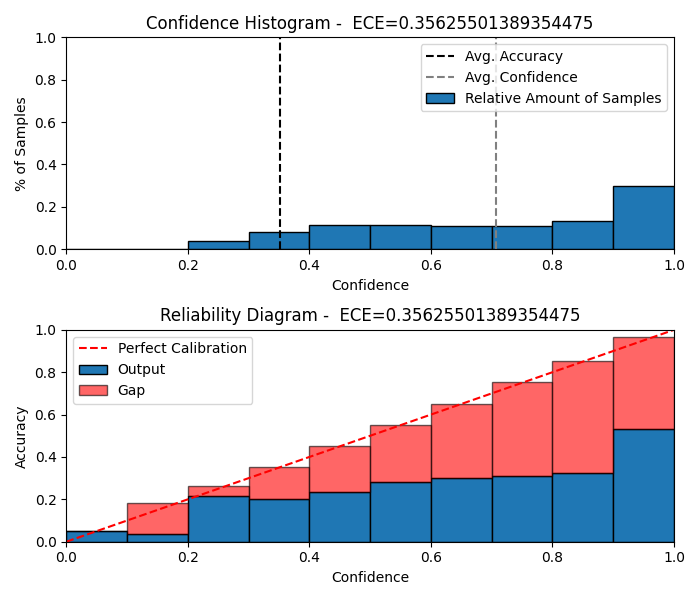}
    \includegraphics[width=0.24\linewidth]{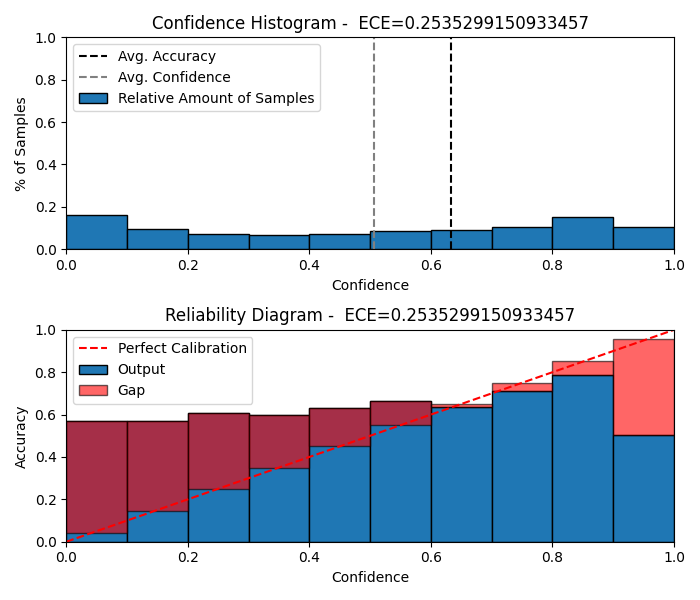}
    \includegraphics[width=0.24\linewidth]{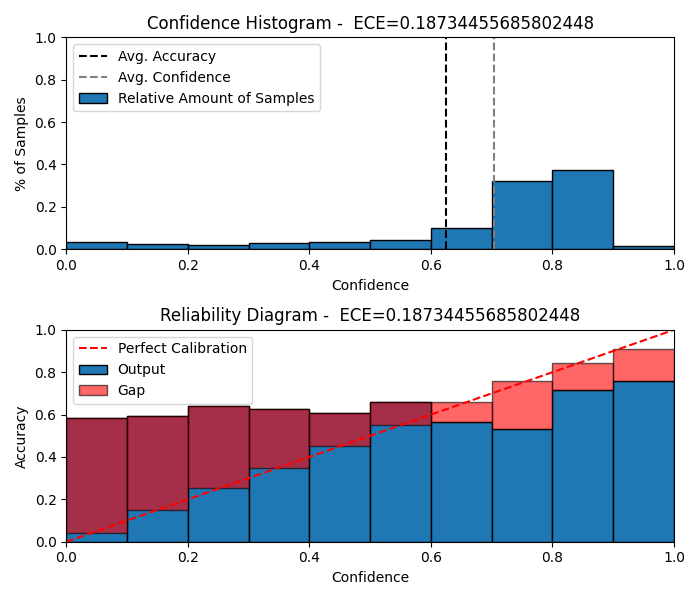}
    \includegraphics[width=0.24\linewidth]{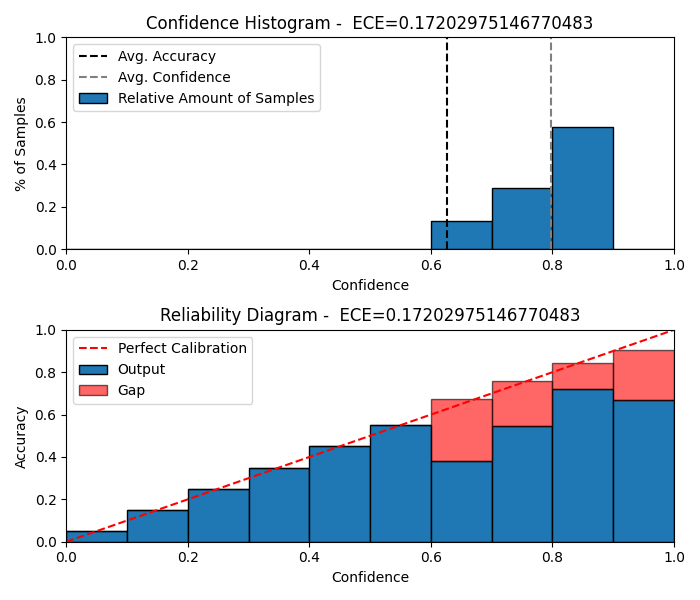}
      \caption{Reliability diagrams for the classifier trained with ERM (leftmost column) versus exit gate calibrations for $E1-E3$ (right hand columns) on \gls*{BiasedMNIST}.}
      \label{fig:bmnist-calibration}
\end{figure}

In Fig.~\ref{fig:coco-calibration} and ~\ref{fig:bmnist-calibration} we show the reliability diagrams for ERM model (leftmost column) and for each exit ($E1-E3$) for OccamResNet for \gls*{COCO} and \gls*{BiasedMNIST} respectively. In terms of model calibration, OccamNet reduces the expected calibration error (ECE) to some extent, yet there is a large room for improvement, which is an interesting direction to pursue.

% \subsection{CAM Visualizations}
% \textbf{Failures cases of \gls*{OccamNets}}
% \begin{itemize}
%     \item Looking at very small regions
% \end{itemize}

\subsection{Other Architectures and Tasks.}
We tested \gls*{OccamNets} implemented with CNNs; however, they may be beneficial to other architectures as well. The ability to exit dynamically could be used with transformers, graph neural networks, and feed-forward networks more generally. There is some evidence already for this on natural language inference tasks, where early exits improved robustness in a transformer architecture~\cite{zhou2020bert}. It would be interesting to evaluate multiple existing early exit mechanisms~\cite{scardapane2020should} for their abilities to discard spurious correlations. Furthermore, adapting the early exit ideas to non-classification tasks e.g., regression may require small changes e.g., exiting based on continuous error, which can be explored in future works.

\end{document}